\documentclass[sigconf]{acmart}

\usepackage{fancyhdr}
\usepackage{url}
\usepackage{microtype}
\usepackage{hyperref}

\usepackage{algorithmic}
\usepackage{graphicx}
\usepackage{textcomp}
\usepackage{xcolor}
\usepackage{subcaption}
\usepackage{float}
\usepackage{mathtools}
\AtBeginDocument{%
  }

\setcopyright{acmcopyright}
\copyrightyear{2023}
\acmYear{2023}
\setcopyright{rightsretained}
\acmConference[MICRO '23]{56th Annual IEEE/ACM International Symposium on Microarchitecture}{October 28-November 1, 2023}{Toronto, ON, Canada}
\acmBooktitle{56th Annual IEEE/ACM International Symposium on Microarchitecture (MICRO '23), October 28-November 1, 2023, Toronto, ON, Canada}
\acmDOI{10.1145/3613424.3623779}
\acmISBN{979-8-4007-0329-4/23/10}

\begin{document}

\title{ADA-GP: Accelerating DNN Training By Adaptive Gradient Prediction}


\author{Vahid Janfaza, Shantanu Mandal, Farabi Mahmud, Abdullah Muzahid}
\affiliation{%
  \institution{Texas A\&M University}
  \streetaddress{L.F. Peterson Building, 435 Nagle St}
  \city{College Station}
  \country{USA}
  }
\email{{vahidjanfaza, shanto, farabi, abdullah.muzahid}@tamu.edu}

\renewcommand{\shortauthors}{Janfaza et al.}

\newcommand{\scheme}{ADA-GP}
\newcommand{\am}[1]{\textcolor{red}{AM: #1}}
\newcommand{\vj}[1]{\textcolor{purple}{VJ: #1}}
\newcommand{\sm}[1]{\textcolor{blue}{SM: #1}}
\newcommand{\fm}[1]{\textcolor{green}{FM: #1}}
\newcommand{\aj}[1]{{#1}}
\newcommand{\rb}[1]{\textcolor{red}}
\newcommand{\phasebp}{{Phase BP}}
\newcommand{\phasegp}{{Phase GP}}
\newcommand{\synmodel}{{predictor}}
\newcommand{\syngrad}{{predicted}}

\begin{abstract}
Neural network training is inherently sequential 
where the layers finish the forward propagation in succession, followed by the calculation and back-propagation of gradients (based on a loss function) starting from the last layer. The sequential computations significantly slow down neural network training, especially the deeper ones. Prediction has been successfully used in many areas of computer architecture to speed up sequential processing. 
Therefore,
we propose \scheme, 
which uses gradient prediction 
adaptively to speed up deep neural network (DNN) training while maintaining accuracy. \scheme\ works by incorporating a small neural network to predict gradients for different layers of a DNN model. \scheme\ uses a novel tensor reorganization method 
to make it feasible to predict a large number of gradients. \scheme\ alternates between DNN training using backpropagated gradients and DNN training using predicted gradients. \scheme\ adaptively adjusts when and for how long gradient prediction is used to strike a balance between accuracy and performance. 
Last but not least, we provide a detailed hardware extension in a typical DNN accelerator
to realize the speed up potential from gradient prediction. Our extensive experiments with fifteen DNN models show that \scheme\ can achieve an average speed up of $1.47\times$ with similar or even higher accuracy than the baseline models. Moreover, it consumes, on average, 34\% less energy due to reduced off-chip memory accesses compared to the baseline accelerator.


\end{abstract}

\ccsdesc[300]{Computing methodologies~Neural networks}
\ccsdesc[300]{Computer systems organization~Neural networks}

\keywords{Hardware accelerators, Training, Prediction, Systolic arrays}


\maketitle

\section{Introduction}
\label{sec:intro}
Deep neural networks (DNNs) have shown remarkable success in recent years. They can solve various complex tasks such as recognizing images~\cite{alexnet}, translating languages~\cite{gpt, transformer}, driving cars autonomously~\cite{autonomous_driving}, generating images/texts~\cite{image-generate}, playing games~\cite{alphago}, etc. DNNs achieve their incredible problem solving ability by training on a vast amount of input data. The de facto standard for DNN training is the backpropagation algorithm~\cite{LeCuBottOrrMull9812}. 
This algorithm works by processing input data using the forward pass through the DNN model starting from the first layer to the last. The last layer computes a pre-defined loss function. Then, gradients are calculated based on the loss function and propagated back from the last layer to the first, updating each layer's weights. Thus, the backpropagation algorithm is inherently sequential: a layer's weights cannot be updated until all the layers finish the forward pass and gradients are propagated back to that layer. 
%
%
\begin{figure}[htpb]
\centering
\begin{subfigure}{0.48\columnwidth}
\centering
\includegraphics[width=0.65\columnwidth]{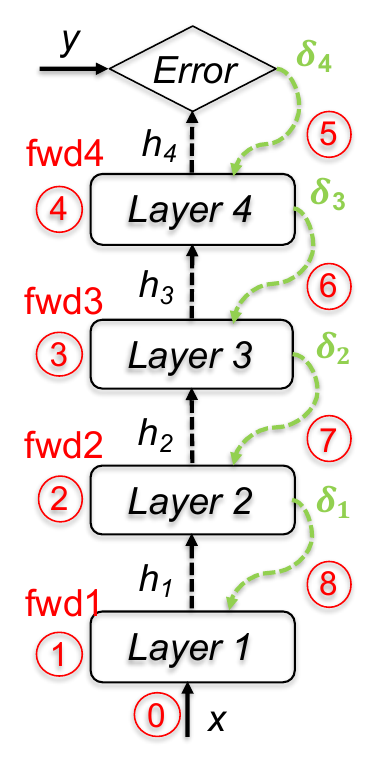}
 \caption{Sequential steps of DNN training algorithm.}
\label{fig-bp}
\end{subfigure}
\hfill
\begin{subfigure}{0.48\columnwidth}
\centering
\includegraphics[width=\columnwidth]{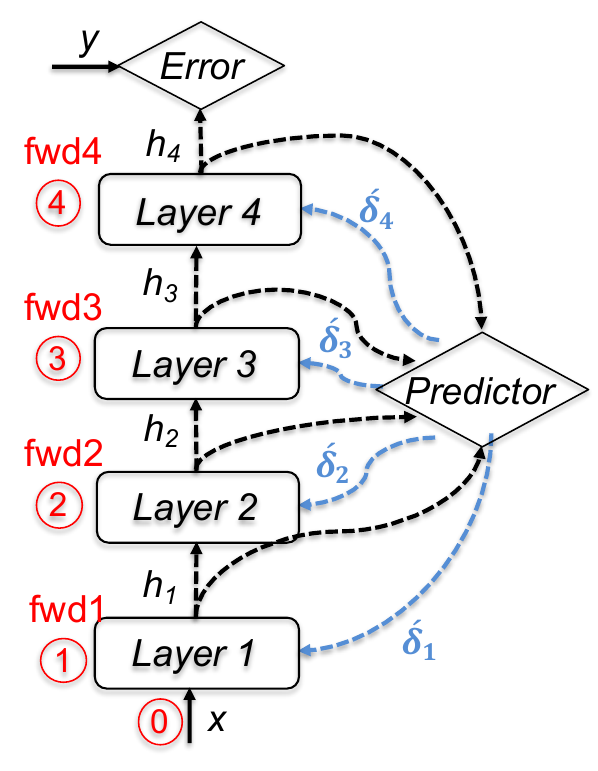}
\caption{DNN training with gradient prediction.}
\label{fig-grad-pred}
\end{subfigure}
\caption{How gradient prediction speeds up DNN training.}
\label{fig-difference}
\end{figure}
This is shown in Figure~\ref{fig-bp} for an example 4-layer model where the gradients $\delta_1, ... \delta_4$ are used 
to update the weights. 

The sequential nature of the backpropagation algorithm makes DNN training a time consuming task. For decades, computer architects have been using {\em prediction} to speed up various processing tasks, including sequential ones. For example, predicting branches, memory dependencies, memory access patterns, synchronizations, etc. have been used in various processor architectures to improve performance. Inspired by this line of research, we set out to investigate whether it is possible to use {\em gradient prediction} to relax the sequential constraints of DNN training. 

There are two major challenges towards gradient prediction.
\begin{enumerate}
    \item {\em The Curse of Scale}: Scalability of gradient prediction arises from two aspects of a DNN model. {\em First,} the number of layers of any recent DNN model can be in the range of hundreds; therefore, having one predictor for each layer is not 
    feasible. {\em Second,} for many layers, the number of gradients (which should be equal to the number of weights 
    in a given layer) is quite large. In some cases, this number can exceed the number of output activations of a layer. 
    Consequently, predicting a large number of gradients for a layer can be challenging.
    
    \item {\em Accuracy vs. Performance}: Always using gradient prediction will speed up DNN training significantly (almost $3x$ speed up by completely eliminating the backpropagation step) but can severely degrade the prediction accuracy. However, if a scheme focuses on predicting 
    high quality gradients and uses them infrequently, it will not affect the prediction accuracy of the DNN model, but reduce the speed up. 
    Therefore, the scheme needs to decide adaptively when and how long to use gradient prediction during DNN training.
\end{enumerate}

To address these challenges, we propose \scheme, the {\bf first} scheme to use gradient prediction for speeding up DNN training while maintaining accuracy. \scheme\ works by incorporating a small neural network model, called a {\em Predictor Model}, to predict gradients of various layers of a DNN model. \scheme\ uses a single \synmodel\ model for all layers. The model takes the output activations of a layer as inputs and predicts the gradients for that layer (as shown in Figure~\ref{fig-grad-pred}). To predict a large number of gradients, \scheme\ uses tensor reorganization (details in Section~\ref{sec-tensor-org}) within 
a batch of input data. When training starts for a DNN model, the weights of the DNN model are initialized randomly. Therefore, the gradients for the first few training epochs (an epoch is defined as one iteration of DNN training using the entire dataset) are more or less random. 
Because of this, \scheme\ uses the standard backpropagation algorithm 
to train the DNN model for a few (e.g., 10) initial epochs. During 
these epochs, \scheme\ trains the \synmodel\ model with the true gradients produced by the backpropagation algorithm. \scheme\ trains the \synmodel\ model with each layer's gradients. After 
the initial epochs, \scheme\ alternates between DNN training using backpropagated gradients and DNN training using gradients predicted by the \synmodel\ model. In other words, for a number of batches (say, $m$), \scheme\ trains the DNN model using the backpropagation algorithm as it is while training the \synmodel\ model with true gradients: we call this {\em \phasebp}. Then, for the next few batches (say, $k$), \scheme\ switches to DNN training using the \synmodel\ model generated gradients: we call this {\em \phasegp}. During \phasegp, the backpropagation algorithm is completely skipped, leading to 
accelerated training of the DNN model. Thus, \scheme\ alternates between \phasebp\ and \phasegp\ gradually adjusting the value of $m$ and $k$ to balance accuracy and performance. 
Finally, we propose some hardware extension in a typical DNN accelerator to implement \scheme\ and realize its full potential.

It should be noted that predicting gradients artificially (as opposed to using backpropagation algorithm) is not new. Several prior works investigate the possibility of utilizing synthetic gradients~\cite{lillicrap2016random, nokland2016direct, xu2017symmetric, balduzzi2015kickback, jaderberg2017decoupled, czarnecki2017understanding, czarnecki2017sobolev, miyato2017synthetic}. This line of work is inspired by the biological learning process and produces synthetic gradients using some form of either controlled randomization or per-layer predictors. However, all of the techniques aim at producing better quality gradients for achieving prediction accuracy and convergence rate at least similar to that of the backpropagation algorithm. {\em  None of the existing techniques investigate synthetic gradients from the performance improvement point of view}. Some techniques~\cite{lillicrap2016random, nokland2016direct} require forward propagation of all layers to finish before synthetic gradients can be produced. The majority of the techniques keep the backpropagation computation as it is and require similar or more training time compared to the backpropagation algorithm alone~\cite{jaderberg2017decoupled, czarnecki2017sobolev, miyato2017synthetic}. Some of the techniques introduce more trainable parameters into the model, leading to an increased training time~\cite{belilovsky2020decoupled}. Last but not least, all of the existing techniques suffer from lower scalability, training stability, and accuracy 
for deeper models. 

\subsection{Contributions}
\label{sec-contributions}
We make the following major contributions:
\begin{enumerate}
    \item \scheme\ is the {\bf first} work that 
    explores the idea of gradient prediction for improving DNN training time. It does so while maintaining the model's accuracy.

    \item \scheme\ uses a single \synmodel\ model to predict gradients for all layers of a DNN model. This reduces the storage and hardware overhead for gradient prediction. 
    Furthermore, \scheme\ uses a {\em novel} tensor reorganization technique among a batch of inputs to predict a large number of gradients.

    \item \scheme\ uses backpropagated and predicted gradients alternatively to balance performance and accuracy. Moreover, \scheme\ adaptively adjusts when and for how long gradient prediction should be used. Thanks to this {\em novel} adaptive 
    algorithm, \scheme\ is able to achieve both high accuracy and performance even for 
    larger DNN models with 
    more sizeable datasets, such as ImageNet.

    \item We propose three possible extensions in a typical DNN accelerator with varying degrees of resource requirements to realize the full potential of \scheme. 
    Additionally, we show how \scheme\ can be utilized in a multi-chip environment with different parallelization techniques to further improve the performance gain.

    \item We implemented \scheme\ in both FPGA and ASIC-style accelerators
    and experimented with {\em fifteen} DNN models using three different datasets - CIFAR10, CIFAR100, and ImageNet. Our results indicate that \scheme\
    can achieve an average speed up of $1.47\times$ with similar or even higher accuracy than the baseline models.
    Also, due to the reduced off-chip memory accesses during the weight updates using predicted gradients, 
    \scheme\ consumes 34\% less energy compared to the baseline accelerator.
\end{enumerate}

\section{Related Work}
\label{sec-back}

\begin{figure*}[!h]
\begin{subfigure}{0.5\linewidth}
    \centering
    \includegraphics[width=0.9\columnwidth]{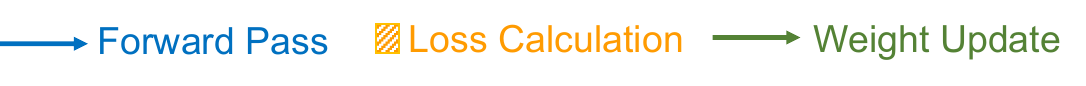}
    \label{fig-label-training}
\end{subfigure}
\vspace{-0.4cm}
\end{figure*}

\begin{figure*}[!h]
\centering
\begin{subfigure}{0.24\linewidth}
    \centering
    \includegraphics[width=0.95\columnwidth]{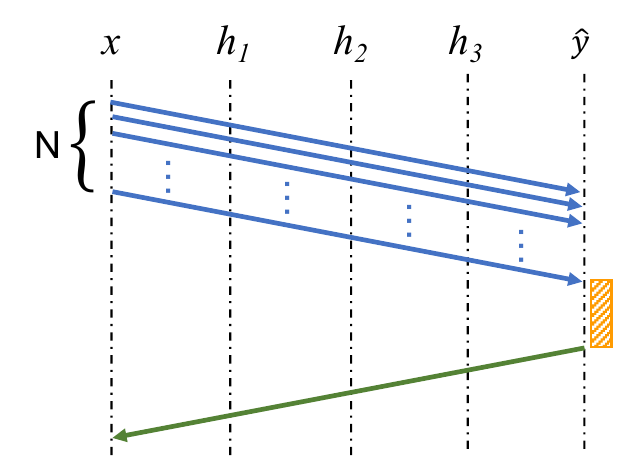}
    \caption{GD}
    \label{fig-sgd}
\end{subfigure}
\hfill
\begin{subfigure}[t]{0.22\linewidth}
    \centering
    \includegraphics[width=0.95\columnwidth]{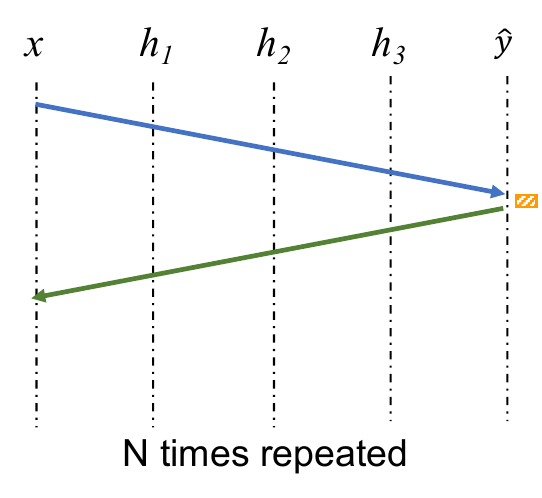}
    \caption{SGD}
    \label{fig-sgd}
\end{subfigure}
\hfill
\begin{subfigure}[t]{0.24\linewidth}
    \centering
    \includegraphics[width=0.95\columnwidth]{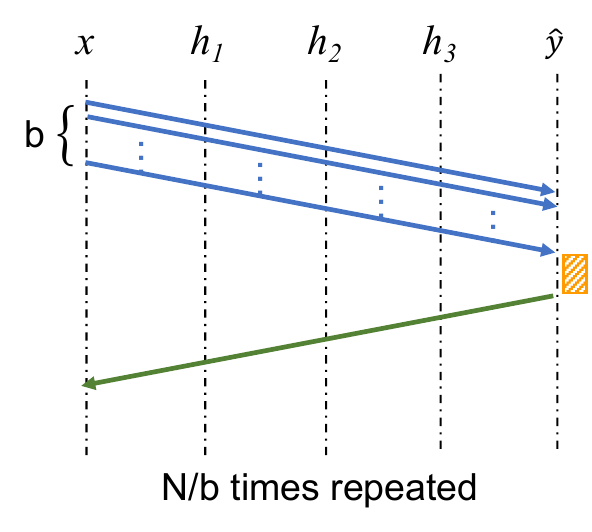}
    \caption{MBGD}
    \label{fig-mbgd}
\end{subfigure}
\hfill
\begin{subfigure}[t]{0.28\linewidth}
    \centering
    \includegraphics[width=\columnwidth]{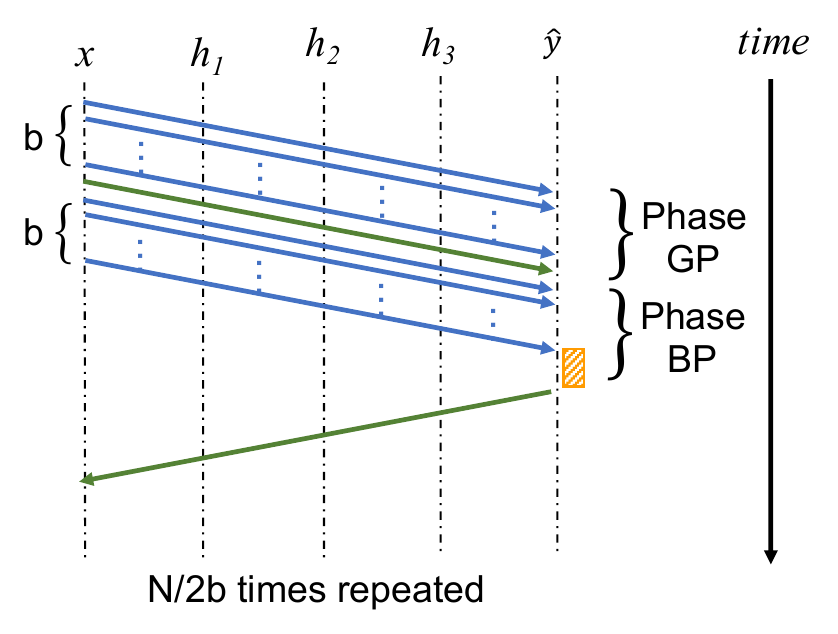}
    \caption{\scheme\ (this work) }
    \label{fig-ada-gp-learning}
\end{subfigure}
%
\caption{Comparison of the learning process and weight updates in (a) Gradient Descent (GD), (b) Stochastic GD (SGD), (c) Mini-batch GD (MBGD), and (d) \scheme. An arrow represent the direction of computations through different layers of the network. Here, the network has four layers including three hidden layers; $N$ is the size of input dataset and $b$ is the batch size. Loss calculation is shown as proportional to the computation amount.}
\label{fig-ADA-GP-vs-SGD}
\end{figure*}


Training of a neural network is done using many input-label pairs such as $(x, y)$ with $x$ and $y$ being the input and the corresponding desired label. In the forward pass, the prediction $\hat{y}$ is calculated, whereas the backward pass calculates the prediction error (i.e., loss) at the output layer and propagates it back through the earlier layers to calculate the weight gradients relative to the loss. The weight gradients are used to update the weights of the network. In case of the Gradient Descent (GD) algorithm, all inputs of the training dataset are used to calculate a single loss and perform a single iteration of weight update. Therefore, for larger datasets, GD becomes painstakingly slow. An alternative is Stochastic GD (SGD) where a single input is randomly chosen from the entire dataset to calculate the loss and perform a single iteration of weight update. For larger datasets, SGD is faster but suffers from lower prediction accuracy. A commonly used middle ground is called Mini-batch GD (MBGD) where a batch of random inputs from the dataset is used to calculate the loss and perform a single iteration of weight update. Whether it is GD, SGD, or MBGD, weight update is always dependent on loss calculation which is dependent on processing the input through the forward pass. The only difference among these approaches is the number of inputs that need to be processed. \scheme\ is fundamentally different from these approaches because (in Phase GP) it allows weight updates to be done in parallel with the froward pass without requiring any loss. This is shown in Figure~\ref{fig-ADA-GP-vs-SGD}.


Jaderberg et al.~\cite{jaderberg2017decoupled} proposed Decoupled Neural Interface (DNI) where a layer receives synthetic gradients from an auxiliary model after output activations of the layer are calculated. The predicted gradients can be used to update the weights of the layer. The auxiliary model is trained based on the backpropagated gradients and the predicted gradients. In other words, DNI requires the backpropagation algorithm to proceed as usual. When a layer has the backpropagated gradients available, these gradients are compared against the predicted gradients, and the auxiliary model is updated. Thus, DNI does not eliminate the backpropagation step at all. Instead, it increases  computations of the backpropagation step by including the auxiliary model update as part of the backpropagation step. That is why, DNI does not improve training time. In fact, it slows down the training time. {\em This is different from \scheme, where the backpropagation step is adaptively skipped as the DNN training proceeds}. Speed up of \scheme\ comes from skipping the backpropagation step altogether. 
Moreover, the DNI approach was shown to work only for small networks (up to 6 layers) and small datasets such as MNIST. 
Czarnecki et al.~\cite{czarnecki2017sobolev} explored the benefits of including derivatives in the learning process of the auxiliary model. The proposed method, called Sobolev Training (ST), considers both the second-order derivatives as well as the backpropagated gradients to train the auxiliary model. The intuition is that by including the derivatives, the auxiliary model will produce higher-quality gradients compared to the DNI approach. However, similar to DNI, it does not eliminate the backpropagation step. Rather, ST increases the backpropagation computations even more by including the computations of the second-order derivatives. Therefore, ST slows down DNN training even further.
Miyato et al.~\cite{miyato2017synthetic} proposed a
virtual forward-backward network (VFBN) 
to simulate the actual sub-network above a DNN layer to generate gradients with respect to the weights of that layer. Thus, VFBN does not eliminate backpropagation at all. Instead, it introduces the backpropagation of a different network, namely VFBN. Using this approach, the authors showed comparative accuracy similar to the baseline model with backpropagation-based learning. However, similar to prior approaches, VFBN does not reduce DNN training time.

There is a number of work that uses some form of random or direct gradients from the last layer. Achieving biological plausibility serves as a key motivation for these techniques~\cite{lillicrap2016random, nokland2016direct, xu2017symmetric, balduzzi2015kickback}, which target the removal of weight symmetry and potential gradient propagation in the backward pass. By substituting symmetrical weights with random ones, Feedback Alignment (FA)~\cite{lillicrap2016random} achieves weight symmetry elimination. 
Direct FA~\cite{nokland2016direct} replaces the backpropagation algorithm with random projection, possibly enabling concurrent updates of all layers. The study by Balduzzi et al.~\cite{balduzzi2015kickback} disrupts local dependencies across consecutive layers, allowing direct error information reception by all hidden layers from the output layer.
All of these approaches end up using poor-quality gradients. Consequently, they degrade the prediction accuracy of the DNN model significantly (especially for deeper models) and eventually, end up taking more time to reach the target accuracy.  
Decoupled Greedy Learning~\cite{belilovsky2020decoupled}, Decoupled Parallel Backpropagation (DDG)~\cite{huo2018decoupled}, and Fully Decoupled Training scheme (FDG)~\cite{zhuang2021fully} are other strategies that aim to address sequential dependencies of DNN training. While DDG and FDG have been shown to reduce total computation time, they incur large memory overhead due to the storage of a large number of intermediate results. Moreover, they also suffer from weight staleness. 
Feature Replay (FR) ~\cite{huo2018training} similarly breaks backward dependency by recomputation. Its performance has been shown to surpass that of backpropagation in various deep architectures. However, FR has a greater computation demand, leading to a slower training time compared to DDG.
Finally, these works require all layers to finish the forward propagation first before the weights can be updated. This is different from \scheme\ where weights of a layer can be updated as soon as the output activations are calculated. \scheme\ does not need to wait for the forward propagation of all layers to finish.

There are a number of parallelization strategies for DNN training. 
Data Parallelism ~\cite{goyal2017accurate, li2020taming, sergeev2018horovod, you2018imagenet} is a widespread method for scaling up the training processes on parallel machines.
However, this method encounters efficiency challenges due to gradient synchronization and model size. Operator Parallelism offers a solution for training larger models by dividing layer operators among multiple workers but faces higher communication requirements. Hybrid techniques ~\cite{jia2019beyond, krizhevsky2014one} combining operator and data parallelism also encounter similar issues. Pipeline Parallelism ~\cite{huang2019gpipe, fan2021dapple, li2021chimera, jain2020gems} has been extensively explored to reduce communication volume by partitioning the model in layers, assigning workers to pipeline the layers, and processing micro-batches sequentially. 
However, \scheme\ is orthogonal to this line of work and can be applied in conjunction with any of these approaches.

\section{Adaptive Gradient Prediction}
\label{sec:idea}

\begin{figure}[htpb]
\centering
\includegraphics[width=0.9\columnwidth]{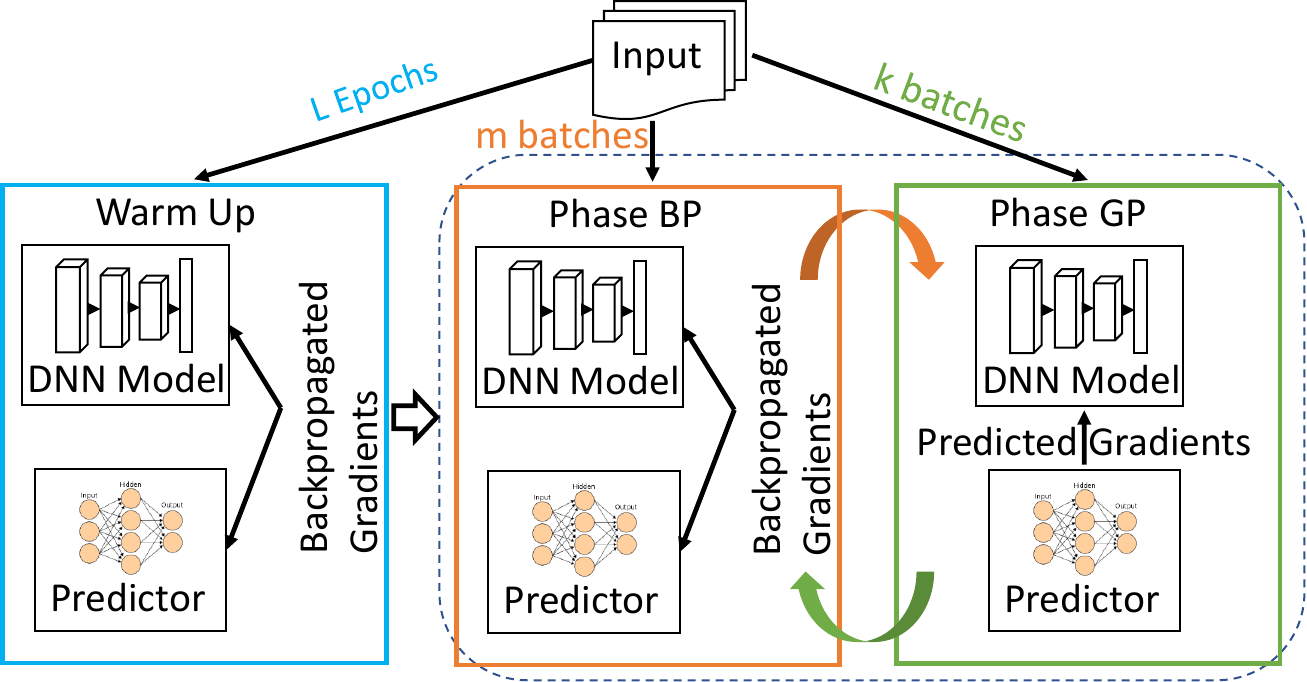}
\caption{Overview of how \scheme\ uses gradient prediction for DNN training.}
\label{fig-overview}
\end{figure}

\subsection{Overview}
\label{sec-overview}
\scheme\ works in three phases. When DNN training starts, the DNN model is initialized and trained using the standard backpropagation algorithm. 
During the first few epochs (e.g., $L$ epochs), the predictor model is trained with the true (backpropagated) gradients without using any of the \syngrad\ gradients in model training. This is the {\em Warm Up} phase (this is reminiscent of the warm up step used in the micro-architectural simulation). Keep in mind that an epoch is one iteration of DNN training with the entire input dataset. 
Afterward, \scheme\ alternates between the backpropagation (Phase BP) and gradient prediction (Phase GP) phases within an epoch. In Phase BP, the DNN model as well as the \synmodel\ model 
are trained with the backpropagated gradients. This is similar to the Warm Up phase except it lasts for a few (say, $m$) batches of input data during an epoch. Then, \scheme\ starts using the \syngrad\ gradients from the \synmodel\ model while skipping the backpropagation step altogether. The skipping of backpropagation leads to an accelerated training in Phase GP. Phase GP lasts for a few (say, $k$) batches of input data. After that, \scheme\ operates in Phase BP followed by Phase GP mode. This continues with the value of $m$ and $k$ adapting over time as the DNN training progresses. Thus, \scheme\ alternates between learning the actual gradients (from backpropagation) and applying the \syngrad\ gradients (after learning).

\begin{figure*}[!h]
\centering
\begin{subfigure}[t]{0.33\textwidth}
    \centering
    \includegraphics[width=\columnwidth]{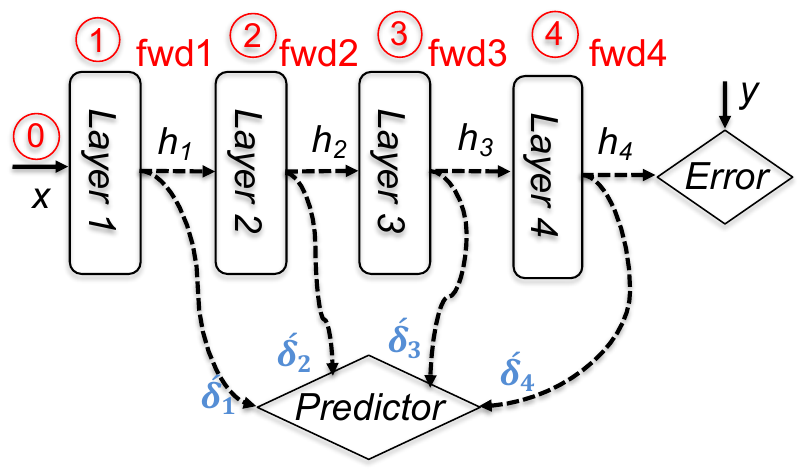}
    \caption{Forward propagation of \scheme\ in \phasebp.}
    \label{ADA-GP-fw-phase-bp}
\end{subfigure}
\hfill
\begin{subfigure}[t]{0.32\textwidth}
    \centering
    \includegraphics[width=\columnwidth]{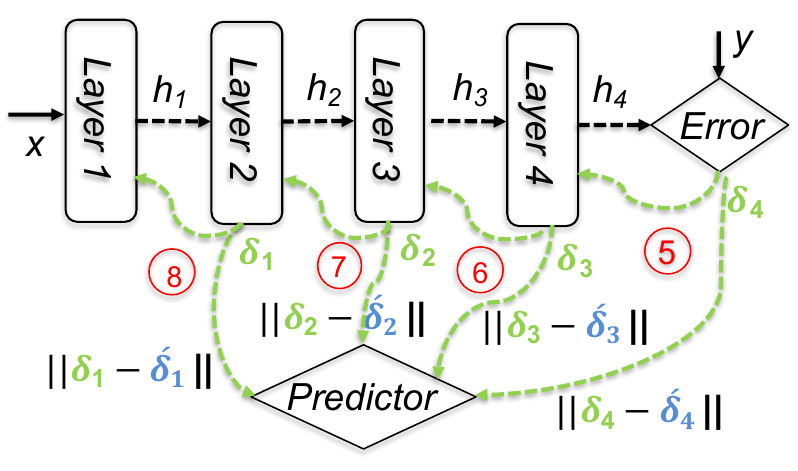}
    \caption{Backpropagation of \scheme\ in \phasebp.}
    \label{ADA-GP-bw-phase-bp}
\end{subfigure}
\hfill
\begin{subfigure}[t]{0.32\textwidth}
    \centering
    \includegraphics[width=\columnwidth]{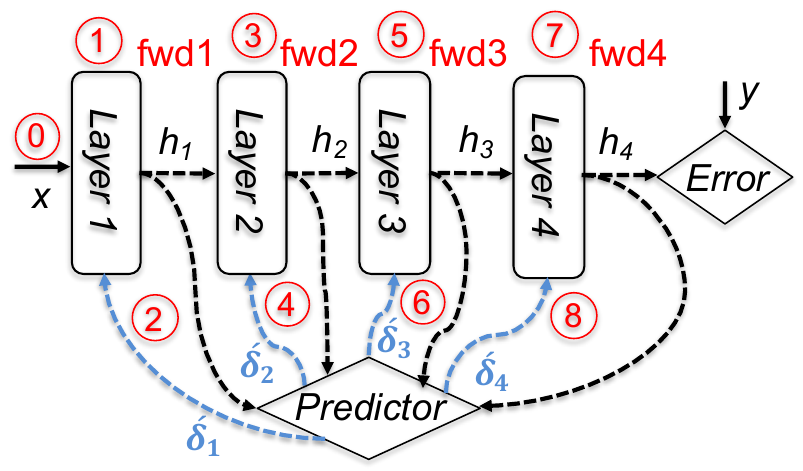}
    \caption{Overall training of \scheme\ in \phasegp. 
    }
    \label{ADA-GP-overall-phase-gp}
\end{subfigure}
%
\caption{The structure of \scheme\ in a) forward propagation of \phasebp, b) backward propagation of \phasebp, and c) comprehensive processes within \phasegp\ that train the initial model using the \syngrad\ gradients by \synmodel\ model.}
\label{fig-ADA-GP-overal-structure}
\end{figure*}

\subsection{Warm Up of \scheme}
\label{sec-main-idea}
The intuition behind the Warm Up phase is to initialize the \synmodel\ model and ramp up its gradient prediction ability.
Since the DNN model is initialized randomly, the backpropagated gradients are more or less random for the initial few epochs. The \synmodel\ model learns from 
the backpropagated gradients of each layer. As a result, the \syngrad\ gradients are even worse in quality during these epochs. 
Because of this, \scheme\ does not apply the \syngrad\ gradients to update the DNN model. Instead, the backpropagated gradients are used for that purpose. Presumably, after few epochs, say $L$, the \synmodel\ starts to produce gradients that are close to the actual backpropagated gradients,
at which point \scheme\ enters into Phase BP and GP.


\subsection{\phasebp\ of \scheme}
\label{sec-phase-bp}
 In \phasebp, both the original and \synmodel\ models are trained based on the true gradients. Contrary to the DNI~\cite{jaderberg2017decoupled} method, which utilizes synthetic gradients for training the original model and true gradients for training the \synmodel\ model, \phasebp\ of \scheme\ calculates the \syngrad\ gradients but does not apply them to the original model's training. Instead, the true gradients are employed for training both the original and \synmodel\ models. This technique maintains a high accuracy for both models while retaining the performance 
 of the DNI approach~\cite{jaderberg2017decoupled}.
Figure~\ref{ADA-GP-fw-phase-bp} \&~\ref{ADA-GP-bw-phase-bp} depict the training approach of Phase BP for a model with 4-layers.
%

%
%
As illustrated in Figure~\ref{ADA-GP-fw-phase-bp}, unlike the DNI approach, the weights of the layers are not updated during the forward propagation (steps \raisebox{.6pt}{\textcircled{\raisebox{-.9pt} {0}}}, \raisebox{.6pt}{\textcircled{\raisebox{-.9pt} {1}}}, \raisebox{.6pt}{\textcircled{\raisebox{-.9pt} {2}}}, \raisebox{.6pt}{\textcircled{\raisebox{-.9pt} {3}}}, and \raisebox{.6pt}{\textcircled{\raisebox{-.9pt} {4}}}) using \syngrad\ gradients. Nevertheless, the \syngrad\ gradients $\delta_1'$, $\delta_2'$, $\delta_3'$, and $\delta_4'$ are still calculated with the \synmodel\ model based on the output activations in each layer. These \syngrad\ gradients are compared against the true gradients (i.e., $\delta_1$, $\delta_2$, $\delta_3$, and $\delta_4$) and the \synmodel\ model is trained during the backward propagation. 
Figure~\ref{ADA-GP-bw-phase-bp} shows the backpropagation in \phasebp.
As shown in Figure~\ref{ADA-GP-bw-phase-bp}, two operations are performed when calculating the true gradients of each layer (steps \raisebox{.6pt}{\textcircled{\raisebox{-.9pt} {5}}}, \raisebox{.6pt}{\textcircled{\raisebox{-.9pt} {6}}}, \raisebox{.6pt}{\textcircled{\raisebox{-.9pt} {7}}}, and \raisebox{.6pt}{\textcircled{\raisebox{-.9pt} {8}}}): 1) the layer weights are updated, and 2) the \synmodel\ model is trained.
As shown in these figures, in \phasebp, the original model undergoes the standard backpropagation step, while the \synmodel\ model is trained concurrently.

\subsection{\phasegp\ of \scheme}
\label{sec-phase-gp}
In \phasegp, the standard backpropagation process is skipped, and the original model is trained based on the \syngrad\ gradients. Furthermore, the \synmodel\ model's training is skipped in this phase.
Figure~\ref{ADA-GP-overall-phase-gp} presents the \scheme\ process in \phasegp. It is important to note that \phasegp\ is applied on a new batch of inputs, following the completion of \phasebp\ with the previous batch. 
As shown in Figure~\ref{ADA-GP-overall-phase-gp}, \phasegp\ does not have the true gradient calculations, and it uses the \syngrad\ gradients to update the original model's weights. Also, in this phase, \scheme\ does not train the \synmodel\ model. 

\subsection{Adaptivity in \scheme}

Following the Warm Up phase, \scheme\ transitions to its
standard operation and adaptively alternates between the two primary phases - Phase BP and GP. Initially, it proceeds with \phasegp, utilizing the \syngrad\ gradients to train the original model. This phase persists for $k$ batches before switching to \phasebp\ for $m$ batches. At the outset, $m < k$. 
This means that, at the beginning, \scheme\ uses predicted gradients more than the true gradients. \scheme\ gradually increases the value of $m$ throughout the training process. As training gets closer to the end, the value of $m$ becomes equal to $k$. From this point onward, the number of training batches in \phasebp\ is equal to that in \phasegp\ until the end of the training, and \scheme\ no longer modifies $m$. The reasoning behind this approach is that the model is mostly random at the beginning and has a certain threshold regarding gradient accuracy. However, during the later epochs, the gradients' changes need to be increasingly precise, necessitating higher quality gradients.

For simplicity in our implementation, we performed some experiments to fix the values of $m$ and $k$ and came up with a simple and efficient heuristic. After the Warm Up phase, we set the $k:m$ ratio to $4:1$ (four batches in phase GP and one batch in phase BP) for the next $4$ epochs. Later, we changed the ratio to $3:1$ for another $4$ epochs. Following this pattern, the ratio was then changed to $2:1$, and ultimately settled at $1:1$ for the remainder of the training.




\subsection{Tensor Reorganization }
\label{sec-tensor-org}
Often the \synmodel\ may need to predict a large number of gradients. For that purpose, \scheme\
rearranges the output activations of a DNN layer prior to forwarding them to the \synmodel\ model. This is done to 1) maintain the \synmodel\ model's compact size, and 2) ensure higher quality for the \syngrad\ gradients. 

The primary challenge in predicting the gradients of weights for each layer lies in the fact that the number of weights in some layers is significantly large. 
When a small \synmodel\ tries to predict a large number of gradients, it not only produces poor-quality gradients but also increases the training time of the \synmodel\ itself. 
For example, consider the fourth layer of the VGG13 model~\cite{simonyan2014very} - Conv2d(in\_channels=128, out\_channels=256, kernel\_size=(3,3), stride=1, padding=1). In this layer, the output activation size is (batch\_size, 256, 28, 28). Consequently, the number of trainable weight-related parameters that the \synmodel\ model should predict is $128\times256\times3\times3$. A simple \synmodel\ model with a single fully connected layer would require an input size of $batch\_size\times256\times28\times28$ and an output size of $128\times256\times3\times3$, necessitating substantial memory storage and computational overhead.

To address this issue, we introduce a {\em novel} tensor reorganization technique. It is based on the observation that every input sample within a batch contributes to the weight update. Thus, by taking the average of output activations across the batch, we account for the combined effect of all samples. Furthermore, each individual output channel can be thought of as a distinct training sample (within a batch) with respect to the predictor. 

\begin{figure}[htpb]
\centering
\includegraphics[width=\columnwidth]{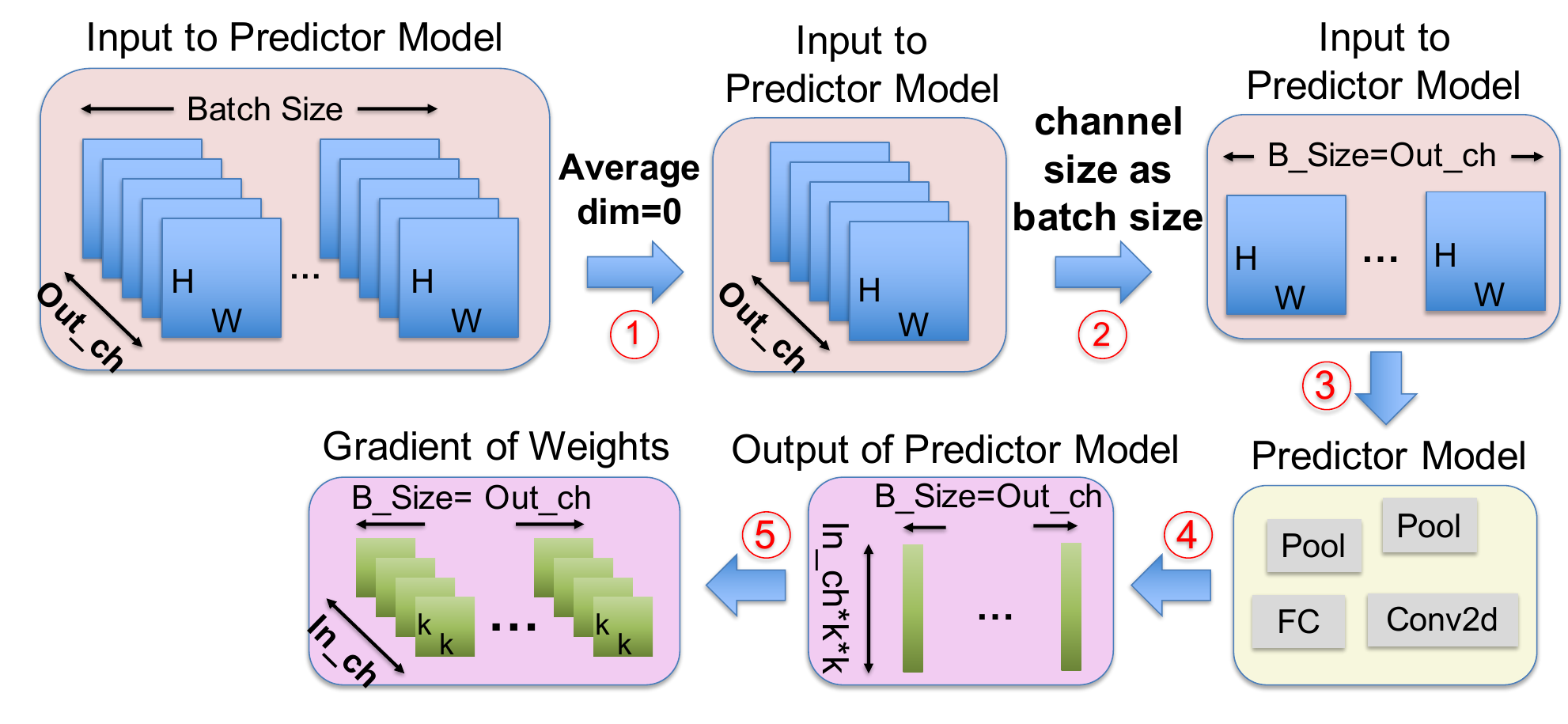}
\caption{An overview of tensor reorganization for generating gradients of a convolution layer, Conv2d(in\_channels=in\_ch, out\_channels=out\_ch, kernel\_size=(k,k)). The \syngrad\ gradient dimensions should match the weight dimensions (in\_ch, out\_ch, k, k). \synmodel\ model input has the shape of (batch\_size, out\_ch, W, H).}
\label{gradient-reorganization}
\end{figure}

Figure~\ref{gradient-reorganization} shows how tensor reorganization works for a convolution layer Conv2d(in\_ch, out\_ch, kernel\_size=(k,k)). In this figure, the output activation size is (batch\_size, out\_ch, W, H), where W and H indicate width and height. This is also the input size to the \synmodel\ model. First, we calculate the average across the batch 
to account for the effects of all samples in a batch, resulting in a tensor with size=(out\_ch, W, H). Considering
that 
each layer's filters 
have 
unique impacts on the output gradients, we can treat out\_ch as the batch size for the \synmodel\ model's input. In our example, we have out\_ch filters with a size of $in\_ch\times k\times k$, generating an output with a channel size of out\_ch, where each filter is individually convoluted with inputs to create a single output. The reshaped tensor (input) for \synmodel\ model becomes (new\_batch\_size=out\_ch, 1, W, H), and the predictor output size becomes (new\_batch\_size=out\_ch,$in\_ch\times k\times k$). 
To generalize the \synmodel\ model for all layers in a large DNN model, we utilize several pooling layers and a small Conv2d layer based on the input size, followed by a single fully connected layer responsible for predicting gradients. Note that the fully connected layer size depends on the largest layer of the DNN model. Therefore, for smaller layers, we simply mask and skip output operations based on the required size.

\subsection{Timeline of \scheme }
\begin{figure}[htpb]
\centering
\includegraphics[width=0.9\columnwidth]{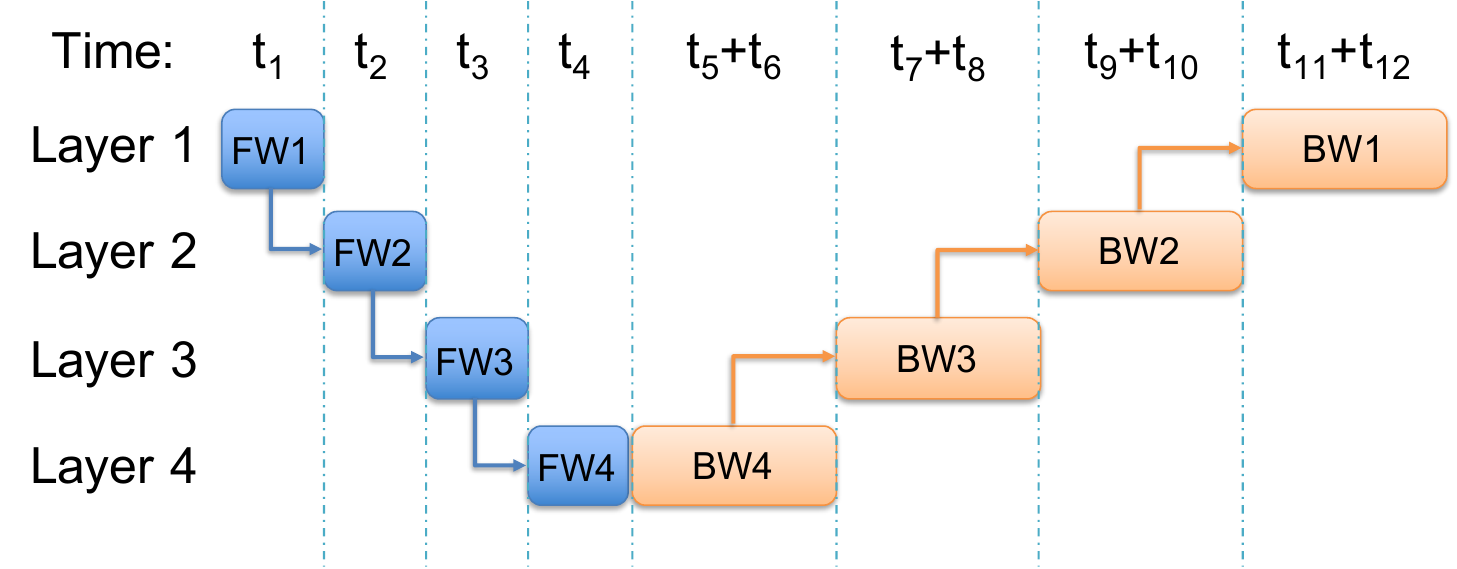}
\caption{Baseline system timeline for a 4-layer neural network model.}
\label{baseline-timeline}
\end{figure}

Figure~\ref{baseline-timeline} illustrates the timeline of the baseline system for a 4-layer neural network model. We assume that the duration of the backward (BW) pass is twice as long as the forward (FW) pass. To simplify the explanation, we focus on the timeline for a single-chip system. We will explore further details about multi-chip pipelining techniques in Section~\ref{Multi-Device-scenario}. As depicted in Figure~\ref{baseline-timeline}, it is evident that the baseline system requires 12 time steps to complete the operation of a 4-layer model for a single batch. In this figure, the duration of each time step is equivalent to the FW pass time for one layer. Throughout the remainder of this section, we will employ this definition of a $step$ in our explanation. 

\begin{figure}[htpb]
\centering
\includegraphics[width=\columnwidth]{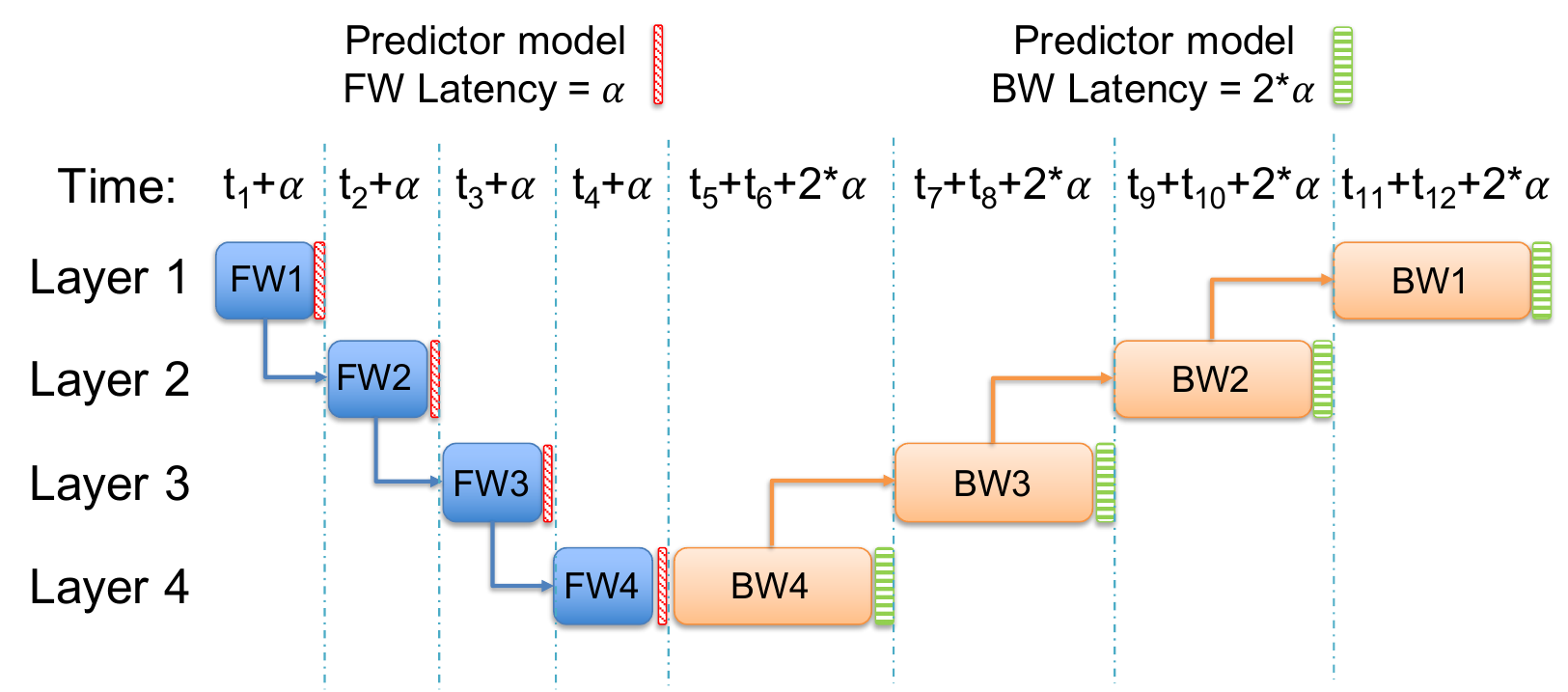}
\caption{\scheme\ timeline in \phasebp.}
\label{ada-gp-timeline-phase-bp}
\end{figure}

Figure~\ref{ada-gp-timeline-phase-bp} shows the timeline of \scheme\ in \phasebp. As mentioned in Section~\ref{sec-phase-bp}, during this phase, \scheme\ trains both the original and \synmodel\ models using the true gradients in the BW pass.
As illustrated in Figure~\ref{ada-gp-timeline-phase-bp}, there is some latency for the FW pass of the \synmodel\ model. This is represented by $\alpha$. This latency is smaller than the FW pass latency of each layer of the original model. Consequently, the latency of the BW pass of the \synmodel\ model is set to 2$\alpha$. As demonstrated in this figure, \scheme\ increases the model's training time by 12$\alpha$. This value is directly linked to the \synmodel\ model's size and the number of operations in its FW and BW pass. 

\begin{figure}[htpb]
\centering
\includegraphics[width=\columnwidth]{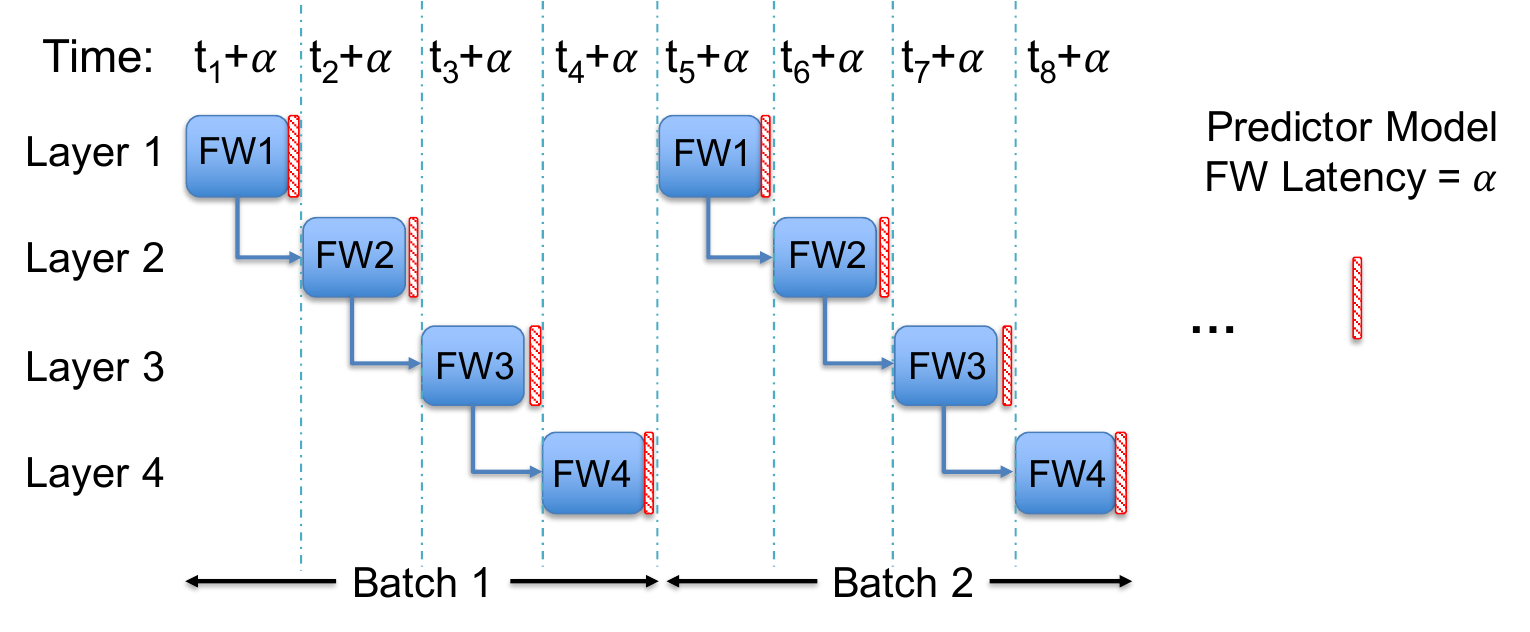}
\caption{\scheme\ timeline in \phasegp.}
\label{ada-gp-timeline-phase-gp}
\end{figure}

\begin{figure*}[!h]
\centering
\begin{subfigure}[b]{0.35\textwidth}
    \centering
    \includegraphics[width=\columnwidth]{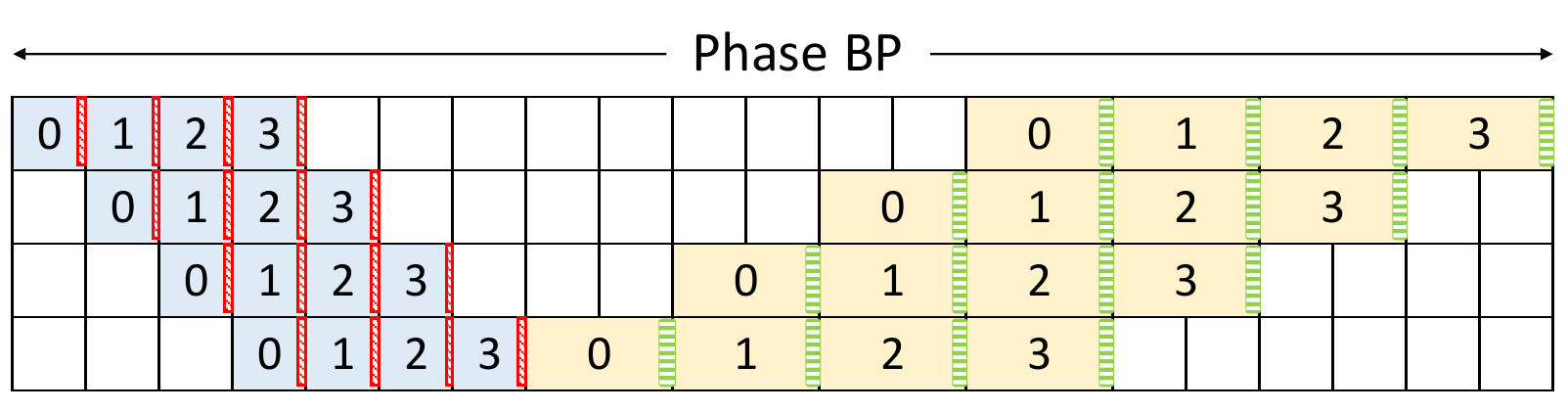}
    \caption{\phasebp}
    \label{GPipe-phase-bp}
\end{subfigure}
\hfill
\begin{subfigure}[b]{0.19\textwidth}
    \centering
    \includegraphics[width=\columnwidth]{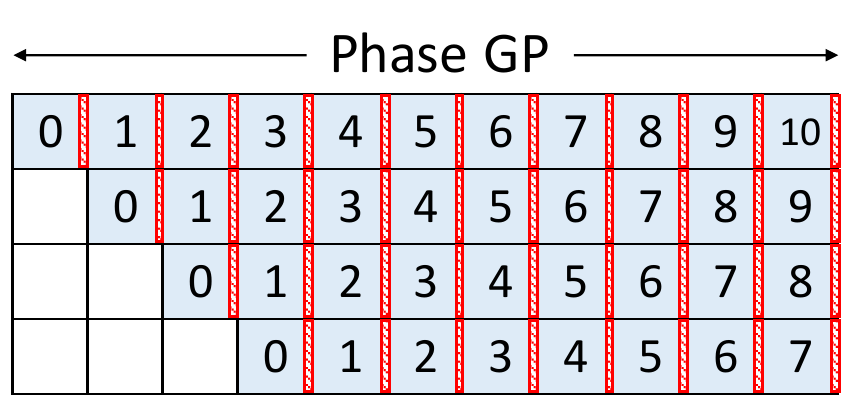}
    \caption{\phasegp}
    \label{GPipe-phase-gp}
\end{subfigure}
\hfill
\begin{subfigure}[b]{0.45\textwidth}
    \centering
    \includegraphics[width=\columnwidth]{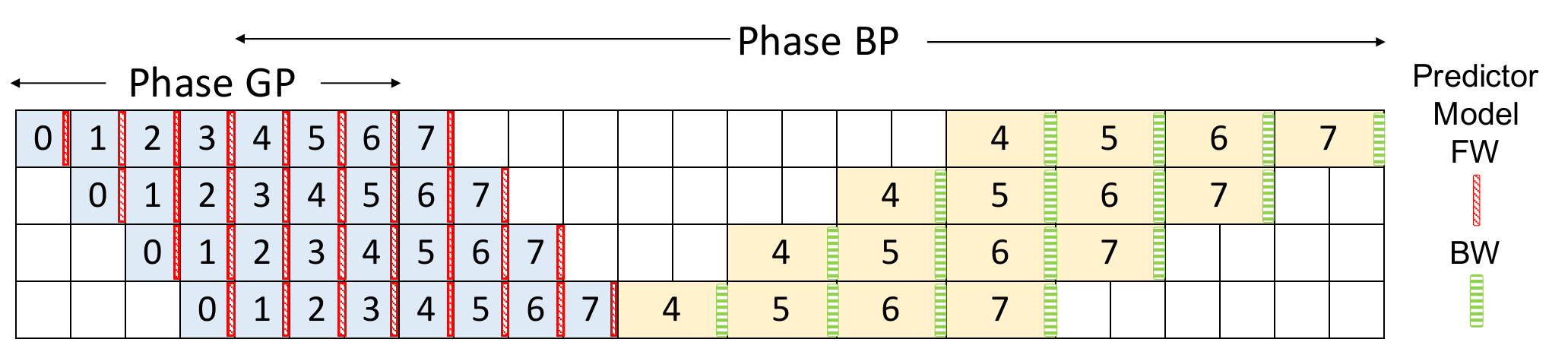}
    \caption{Transition from \phasegp\ to \phasebp}
    \label{GPipe-mix}
\end{subfigure}

\caption{Structure of \scheme\ over GPipe~\cite{huang2019gpipe} during a) \phasebp, b) \phasegp, c) transition from \phasebp\ to \phasegp.}
\label{ADA-GP-GPipe}
\end{figure*}

\begin{figure*}[!h]
\centering
\begin{subfigure}[b]{0.35\textwidth}
    \centering
    \includegraphics[width=\columnwidth]{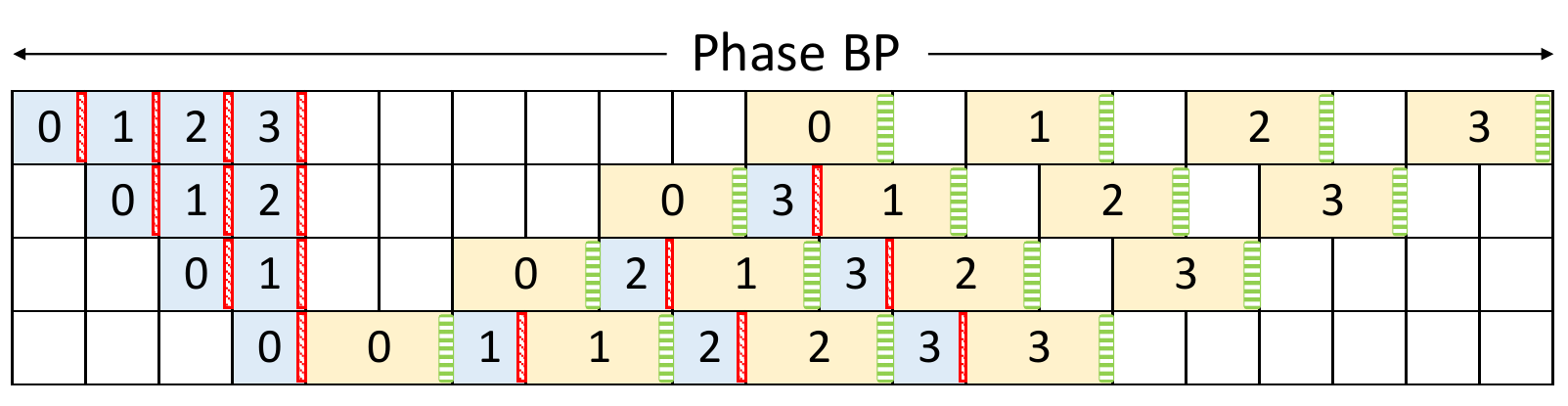}
    \caption{\phasebp}
    \label{Dapple-phase-bp}
\end{subfigure}
\hfill
\begin{subfigure}[b]{0.19\textwidth}
    \centering
    \includegraphics[width=\columnwidth]{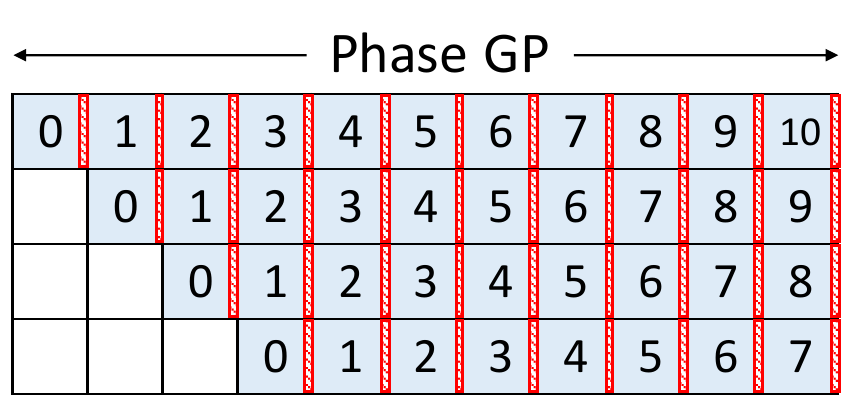}
    \caption{\phasegp}
    \label{Dapple-phase-gp}
\end{subfigure}
\hfill
\begin{subfigure}[b]{0.45\textwidth}
    \centering
    \includegraphics[width=\columnwidth]{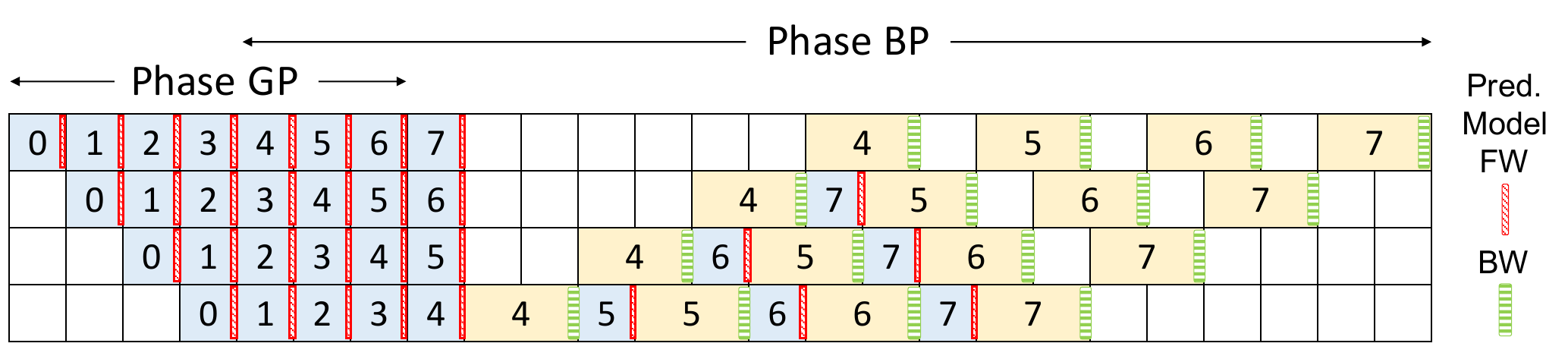}
    \caption{Transition from \phasegp\ to \phasebp}
    \label{Dapple-mix}
\end{subfigure}

\caption{Structure of \scheme\ over DAPPLE~\cite{fan2021dapple} during a) \phasebp, b) \phasegp, c) transition from \phasebp\ to \phasegp.}
\label{ADA-GP-Dapple}
\end{figure*}

\begin{figure*}[!h]
\centering
\begin{subfigure}[b]{0.35\textwidth}
    \centering
    \includegraphics[width=\columnwidth]{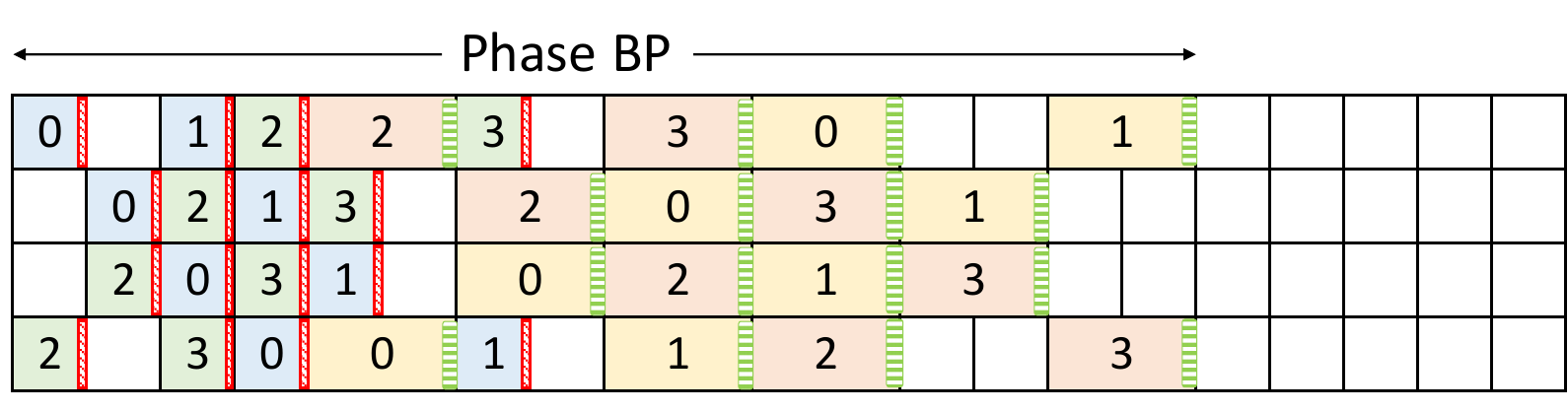}
    \caption{\phasebp}
    \label{Chimera-phase-bp}
\end{subfigure}
\hfill
\begin{subfigure}[b]{0.19\textwidth}
    \centering
    \includegraphics[width=\columnwidth]{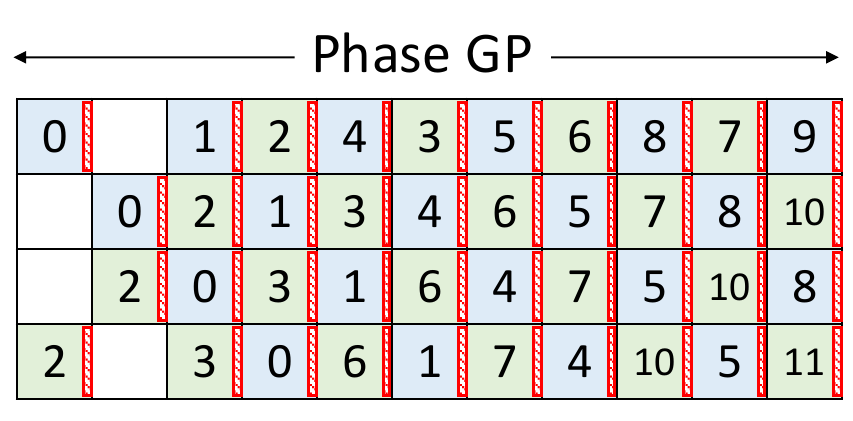}
    \caption{\phasegp}
    \label{Chimera-phase-gp}
\end{subfigure}
\hfill
\begin{subfigure}[b]{0.45\textwidth}
    \centering
    \includegraphics[width=\columnwidth]{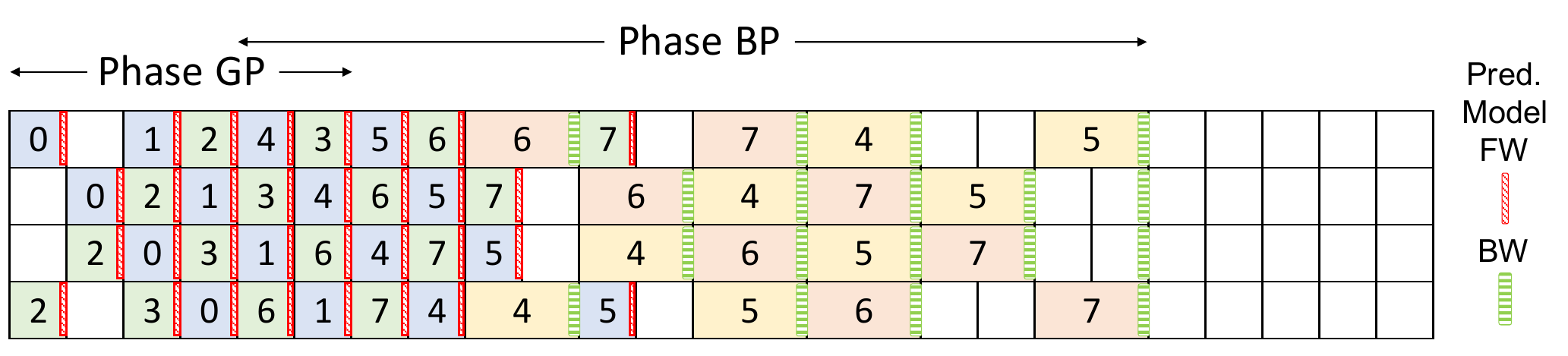}
    \caption{Transition from \phasegp\ to \phasebp}
    \label{Chimera-mix}
\end{subfigure}

\caption{Structure of \scheme\ over Chimera~\cite{li2021chimera} during a) \phasebp, b) \phasegp, c) transition from \phasebp\ to \phasegp.}
\label{ADA-GP-Chimera}
\end{figure*}

Figure~\ref{ada-gp-timeline-phase-gp} presents the timeline of \scheme\ in \phasegp. As mentioned in Section~\ref{sec-phase-gp}, the BW process is skipped in this phase, and the \synmodel\ model is not trained. However, the original DNN model is trained using the \syngrad\ gradients generated by the \synmodel\ model.
In Figure~\ref{ada-gp-timeline-phase-gp}, it is evident that the BW pass is entirely eliminated, leaving only the FW pass of the original model and a minor delay for the FW pass of the \synmodel\ model. Consequently, \scheme\ can minimize the processing time to merely 4+4$\alpha$ steps. As illustrated in Figures~\ref{ADA-GP-fw-phase-bp},~\ref{ADA-GP-bw-phase-bp}, and~\ref{ADA-GP-overall-phase-gp}, \scheme\ is capable of decreasing the processing time for two epochs from 24 steps in the baseline system to 16+16$\alpha$. 
As an added benefit of skipping the BW pass in Phase GP, \scheme\ reduces off-chip traffic. Since the weights are updated as the FW pass proceeds, \scheme\ does not need to load the weights and activations from some off-chip memory as is traditionally done in the case of BW pass. This significantly reduces energy consumption. More details are presented in Section~\ref{energy-consumption-analysis}.

\subsection{\scheme\ in Multi-Device Hardware}
\label{Multi-Device-scenario}
A commonly used approach for accelerating DNN training in multiple devices involve
pipelining techniques to execute several layers concurrently. \scheme\ is orthogonal to this approach and can be integrated with it. To this end, we examine three prominent pipelining strategies - GPipe\cite{huang2019gpipe}, DAPPLE\cite{fan2021dapple}, Chimera\cite{li2021chimera} and explain how \scheme\ can be incorporated with them to further speed up the training process. 
For the ease of explanation, we assume in this section that there are four devices working concurrently, and the batch is divided into four segments, with each device processing one segment at a time.

Figure~\ref{ADA-GP-GPipe} shows how
various \scheme\ phases work when implemented on top of the GPipe approach \cite{huang2019gpipe}. As depicted in Figure~\ref{GPipe-phase-bp}, the \scheme\ operation in \phasebp\ is similar to the original GPipe method. Note that the duration of each step in \scheme\ differs from that in the original GPipe method. The step size of \scheme\ depends on its implementation as outlined in Section~\ref{hardware-implementation}.
Figure~\ref{GPipe-phase-gp} shows the \scheme\ in \phasegp.
In this figure, since \scheme\ eliminates the initial backpropagation process and employs \syngrad\ gradients for weight updates, it can initiate the subsequent batch's process immediately after completing the current batch's forward propagation. In doing so, \scheme\ can fill all gaps present in the original GPipe method. Lastly, Figure ~\ref{GPipe-mix} illustrates how \scheme\ transitions from \phasebp\ to \phasegp\ without causing any additional delay.
Another important point that should be taken into consideration is that \scheme\ reduces the number of synchronization steps to half and can save time and energy due to this reduction.

Figure~\ref{ADA-GP-Dapple} illustrates how various stages of \scheme\ can be integrated with the DAPPLE method~\cite{fan2021dapple}. Like the GPipe strategy, the configuration of \scheme\ in \phasebp\ closely resembles the original DAPPLE design. The depiction of \scheme\ during \phasegp\ can be observed in Figure~\ref{Dapple-phase-gp}. As demonstrated in this figure, \scheme\ effectively eliminates the reliance between forward and backward propagation, filling all gaps in the training procedure. Additionally, Figure~\ref{Dapple-mix} portrays the shift from \phasebp\ to \phasegp.

The complete structure of the various stages of \scheme\ when implemented alongside the Chimera method \cite{li2021chimera} is depicted in Figure ~\ref{ADA-GP-Chimera}. As with earlier strategies, the structure of \scheme\ during \phasebp\ closely mirrors the initial Chimera design. In \phasegp, it is capable of operating all layers concurrently, eliminating any gaps. Furthermore, a transition between phases incurs no additional delays.


\section{Implementation Details}
\label{sec:implement}
\label{sec-impl}

\subsection{Baseline DNN Accelerator}
\begin{figure}[htpb]
\centering
\includegraphics[width=0.8\columnwidth]{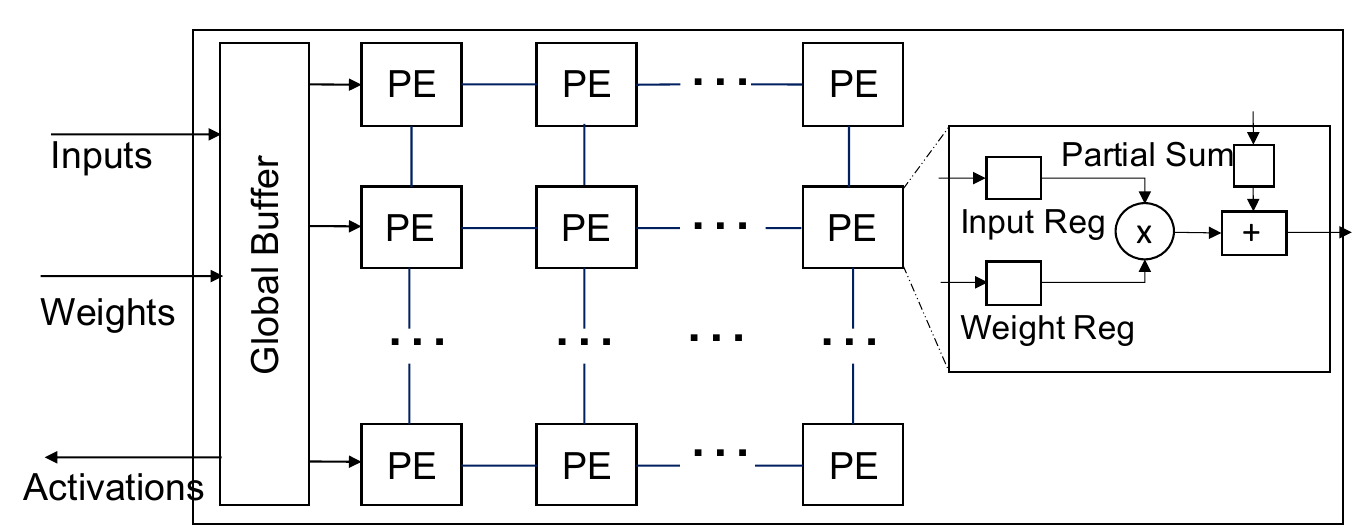}
\caption{Baseline hardware accelerator.}
\label{fig-base}
\end{figure}

Figure~\ref{fig-base} illustrates a standard DNN accelerator design, featuring multiple hardware processing elements (PEs). These PEs are interconnected vertically and horizontally via on-chip networks. A global buffer stores input data, weights, and intermediate results. The accelerator is connected to external memory for inputs and outputs. Each PE is equipped with registers for holding inputs, weights, and partial sums, as well as multiplier and adder units. Inputs and weights are distributed across the PEs, which then generate partial sums following a specific dataflow ~\cite{janfaza2023mercury}.

Various dataflows have been suggested in the literature~\cite{dataflows, eyeriss, maeri, janfaza2023mercury} to enhance different aspects of DNN operations, such as Weight-Stationary (WS) ~\cite{sankaradas2009massively, chakradhar2010dynamically, gokhale2014240}, Output-Stationary (OS)~\cite{du2015shidiannao}, Input-Stationary (IS)~\cite{sze2017efficient}, and Row-Stationary (RS)~\cite{eyeriss}. The dataflow's designation often indicates which data remains constant in the PE during computation.
In the Weight-Stationary (WS) approach, each PE retains a weight in its register, with operations utilizing the same weight assigned to the same PE unit~\cite{dataflows}. Inputs are broadcasted to all PEs over time, and partial results are spatially reduced across the PE array after each time step. This method minimizes energy consumption by reusing filter weights and reducing weight reads from DRAM.
Output-Stationary (OS)~\cite{outputstationary} focuses on accumulating partial results within each PE unit. At every time step, both input and filter weight are broadcasted across the PE array, with partial results calculated and stored locally in each PE's registers. This method minimizes data movement costs by reusing partial results.
Input-Stationary (IS) involves loading input data once and keeping it in the registers throughout the computation. Filter weights are unicasted at each time step, while partial results are spatially reduced across the PE array. This strategy reduces the cost of sequentially reading input data from DRAM.
Row-Stationary dataflow assigns each PE one row of input data to process. Filter weights stream horizontally, inputs stream diagonally, and partial sums are accumulated vertically. Row-Stationary has been proposed in Eyeriss~\cite{eyeriss} and is considered one of the most efficient dataflows to maximize data reuse.

\subsection{\scheme\ Hardware Implementation}
\label{hardware-implementation}

\begin{figure*}[!htpb]
\centering
\begin{subfigure}[t]{0.32\textwidth}
    \centering
    \includegraphics[width=0.85\columnwidth]{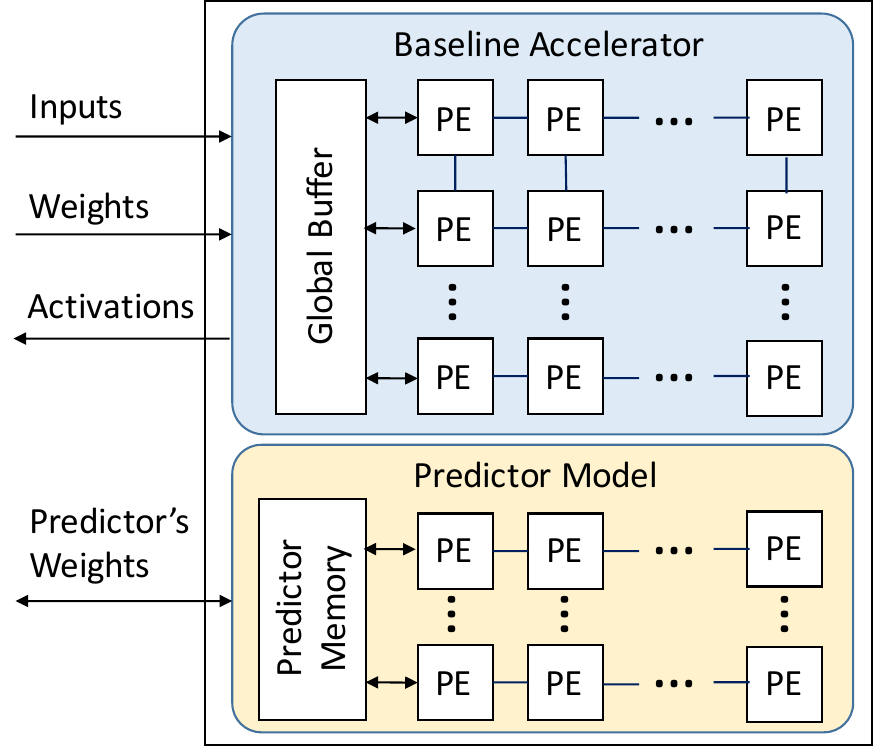}
    \vspace{-0.05cm}
    \caption{\scheme-MAX architecture: This design employs distinct specialized PEs and memory for executing \synmodel\ model calculations. This allows for the simultaneous processing of \synmodel\ model operations alongside the original model.}
    \label{ada-gp-max}
\end{subfigure}
\hfill
\begin{subfigure}[t]{0.32\textwidth}
    \centering
    \includegraphics[width=0.9\columnwidth]{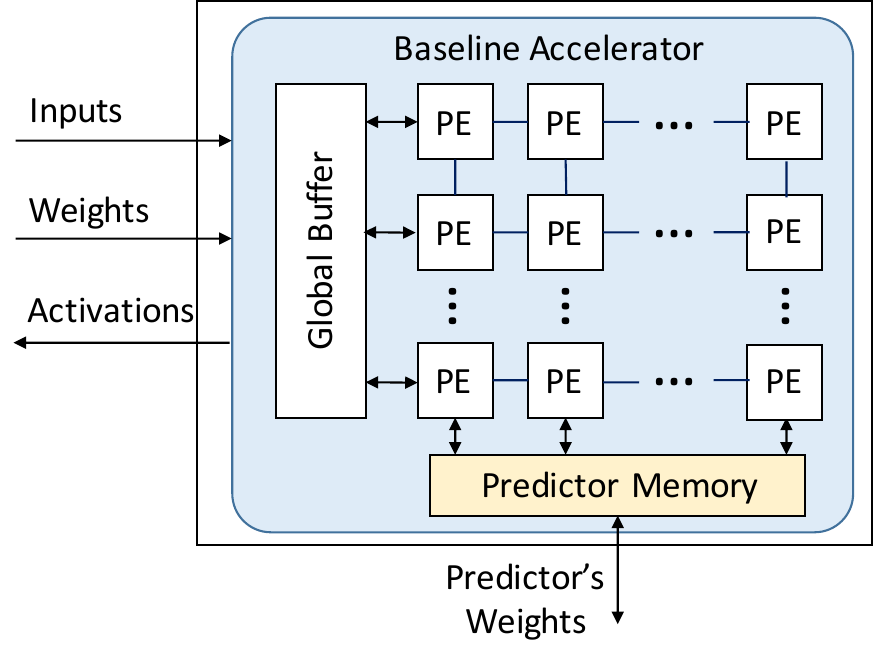}
    \vspace{-0.2cm}
    \caption{\scheme-Efficient architecture: This design does not have a specific PE-Array for \synmodel\ model however it has specific memory for saving the \synmodel's weights and starts the \synmodel\ model computations after finishing the original.}
    \label{ada-gp-efficient}
\end{subfigure}
\hfill
\begin{subfigure}[t]{0.32\textwidth}
    \centering
    \includegraphics[width=0.9\columnwidth]{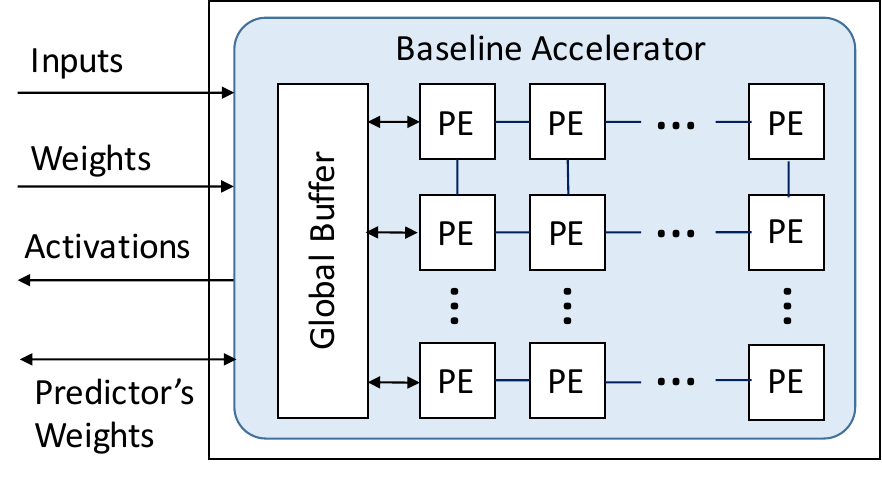}
    \caption{\scheme-LOW architecture: This design has no specific hardware overhead for \synmodel\ model computations. Upon completion of the original model's operation, it initiates the \synmodel\ model's functioning by loading the \synmodel's weights.}
    \label{ada-gp-low}
\end{subfigure}

\caption{Three distinct \scheme\ approaches, balancing the trade-off between hardware overhead and the degree of acceleration. }
\label{ADA-GP-overal-structure}
\end{figure*}

The general architecture of the \scheme\ is similar to the baseline accelerator shown in Figure~\ref{fig-base}. To implement \scheme, we propose three designs, striking a balance between hardware resource constraints and the degree of acceleration. Figure~\ref{ADA-GP-overal-structure} shows the three distinct designs we proposed for \scheme. 

Figure~\ref{ada-gp-max} displays the architecture of \scheme-MAX. This configuration incorporates an additional PE Array and memory for \synmodel\ model computations and weights storage, respectively. Consequently, \scheme\ can initiate the \synmodel\ model's gradient prediction operations concurrently with the original model's computations, overlapping the processes and accelerating training. This design offers the most acceleration but also has more hardware overhead in comparison with other designs.

To offset the hardware overhead of \scheme-MAX, Figure~\ref{ada-gp-efficient} presents the \scheme-Efficient architecture. Instead of an extra PE array for the \synmodel\ model's calculations, this design features a separate memory to store \synmodel\ model's weights and commence its operations immediately after completing the original layer computations. While this configuration saves time and energy consumption related to reading and storing \synmodel's weights, it must wait for the original model's operation to finish before starting the \synmodel\ model computations.

Aiming to further reduce the hardware overhead of the \scheme\ design, Figure~\ref{ada-gp-low} depicts the \scheme-LOW structure. This layout eliminates all additional hardware overhead from the original design and reuses existing resources for \synmodel\ model computations. First, it completes the original model's operations, then, after saving all necessary changes, loads the \synmodel's weights and employs the original PE Array for \synmodel\ model computations and updates. 

\section{Experimental Setup}
\label{sec:experiment}
\subsection{\scheme\ Hardware Setup}
We implemented \scheme\ hardware in both FPGA and ASIC platforms.
For FPGA implementation, we employed the Virtex 7 FPGA board ~\cite{virtex}, configured through the Xilinx Vivado ~\cite{vivado} software. For ASIC implementation, the Synopsys Design Compiler ~\cite{design-compiler} was used, and the design was developed using the Verilog language.
Our implementation utilized a weight stationary accelerator with 180 PEs as the baseline. In the FPGA design, the model's inputs and weights are stored in an external SSD connected to the FPGA. Block memories are employed to load one layer's weights and inputs while storing the corresponding outputs.
Performance, power consumption, hardware utilization, and other hardware-related metrics are gathered from the synthesized and placed and routed FPGA design using Vivado and the synthesized ASIC design using the Design Compiler.

\subsection{\scheme\ Software Setup}
We gathered various model accuracy results from the PyTorch~\cite{pytorch} implementation of \scheme. Fifteen networks are considered, including Inception-V4 \cite{szegedy2016inceptionv4}, Inception-V3 \cite{szegedy2016rethinking}, Densenet201 \cite{huang2017densely}, Densenet169 \cite{huang2017densely}, Densenet161 \cite{huang2017densely}, Densenet121 \cite{huang2017densely}, ResNet152 \cite{he2016deep}, ResNet101 \cite{he2016deep}, ResNet50 \cite{he2016deep}, VGG19 \cite{simonyan2014very}, VGG16 \cite{simonyan2014very}, VGG13 \cite{simonyan2014very}, MobileNet-V2 \cite{howard2018inverted}, Transformer \cite{vaswani2017attention}, and YOLO-v3 \cite{redmon2018yolov3}. Our method is applied to three different datasets: ImageNet \cite{imagenet_cvpr09}, Cifar100 \cite{krizhevsky2009learning}, and Cifar10 \cite{krizhevsky2009learning}. For the transformer, we used the Multi30k \cite{elliott-etal-2016-multi30k} dataset and reported accuracy and BLEU scores \cite{papineni2002bleu} separately in Section \ref{transformer-analysis}. Also, for the YOLO-v3 \cite{redmon2018yolov3} object detection model, we utilized the Pascalvoc \cite{Everingham15} Visual Object Classes dataset.

During the training of \scheme\ and the baseline, the initial learning rate was set to 
0.001
for the original models and 
0.0001
for the \synmodel\ model. We employed SGD with Momentum and Adam optimizers for the original and \synmodel\ models, respectively. Additionally, we utilized the PyTorch ReduceLROnPlateau scheduler with default parameters for adaptive learning rate updates, while a MultiStepLR scheduler was applied for the \synmodel\ model scheduler.
Top 1 accuracy was reported for the various models. To evaluate training costs, end-to-end training costs were calculated.


\section{Evaluation}
\label{sec:result}
Numerous previous studies have explored the potential of employing synthetic gradients in their research ~\cite{lillicrap2016random, nokland2016direct, xu2017symmetric, balduzzi2015kickback, jaderberg2017decoupled, czarnecki2017understanding, czarnecki2017sobolev, miyato2017synthetic}. These approaches generate synthetic gradients through controlled randomization or per-layer predictors. However, none of these methods focus on performance enhancement or skipping the backpropagation step.
Moreover, their accuracy is less \aj{than} or equal to that of backpropagation-based training.
Therefore, at best, those approaches will have accuracy and performance similar to the backpropagation-based training.
\aj{This} is why we use the backpropagation technique as our baseline to compare both the accuracy and performance of \scheme.

\subsection{Accuracy Analysis}
\label{ovr-accuracy}

\subsubsection{Model Accuracy}
We evaluate the accuracy of the proposed method across thirteen different deep-learning models using three distinct datasets: ImageNet, Cifar100, and Cifar10, and compare it with the baseline Backpropagation (BP) approach. 

\begin{table}[!htpb]
\centering
    \caption{Accuracy comparison between \scheme\ and Baseline (BP) for CIFAR10, CIFAR100 \& ImageNet dataset.}
    \label{tbl-acc-merged}
    \scalebox{0.85}{

    \def\arraystretch{1.2}
    \begin{tabular}{||p{1.9cm}||p{0.8cm}|p{0.8cm}||p{0.8cm}|p{0.8cm}||p{0.8cm}|p{0.8cm}||} 
    \hline 
    {} & \multicolumn{2}{c||}{\textbf{CIFAR10}}  &  \multicolumn{2}{c||}{\textbf{CIFAR100}} &  \multicolumn{2}{c||}{\textbf{ImageNet}}  \\ 
     \hline 
     
     & \textbf{BP} &  \textbf{\scheme} &
     \textbf{BP} &  \textbf{\scheme} & \textbf{BP} &  \textbf{\scheme}\\
     \hline\hline
	ResNet50		&	92.97	&	\textbf{93.76} & 75.44  & \textbf{75.73} 	& 74.73  &  73.97 \\\hline
	ResNet101		&	92.78	&	\textbf{93.62} &  73.23 & \textbf{75.38} 	& 76.26  &	75.71 \\\hline
	ResNet152		&	92.8	&	\textbf{93.12} &  72.01 & \textbf{73.7 } 	& 76.68  &	76.23 \\\hline
	Inception-V4	&	91.22	&	\textbf{91.35} &  70.52 & \textbf{72.42} 	& 76.23  &	75.9 \\\hline
	Inception-V3	&	93.04	&	\textbf{93.88} &  76.41 & \textbf{77.68} 	& 74.44  &	73.87 \\\hline
	VGG13			&	91.52	&	\textbf{92.55} & 70.48  & 70.41          	& 70.68  &	\textbf{70.68}  \\\hline
	VGG16			&	91.32	&	\textbf{92.34} & 70.36  & \textbf{70.47} 	& 72.07  &	71.98 \\\hline
	VGG19			&	91.18	&	\textbf{92.51} & 69.69  & \textbf{69.85} 	& 72.94  &	72.83 \\\hline
	DenseNet121		&	93.2	&	\textbf{93.63} &  76.25 & 76.12          	& 75.25  &	74.51 \\\hline
	DenseNet161		&	93.48	&	\textbf{94.19} &  76.87 & \textbf{77.38} 	& 76.43  &	\textbf{76.71} \\\hline
	DenseNet169		&	93.24	&	\textbf{94.15} &  75.57 & \textbf{76.4 } 	& 75.36  &	75.3 \\\hline
	DenseNet201		&	93.26	&	\textbf{94.13} &  76.37 & \textbf{76.96} 	& 75.52  &	\textbf{75.56} \\\hline
	MobileNet	    &	90.08	&	\textbf{91.34} &  68.11 & \textbf{68.47} 	& 69.88  &	69.23 \\\hline

    \end{tabular}}
\end{table}

%

Table~\ref{tbl-acc-merged} presents the accuracy comparison between the proposed \scheme\ and \aj{the} baseline BP for the Cifar10, Cifar100, and ImageNet datasets. As shown in this table, in the Cifar10 dataset, \scheme\ effectively boosted the accuracy of all models by as much as 1.45\% and an average of 0.75\%. When applied to the Cifar100 dataset, \scheme\ similarly yielded improvements, with accuracy enhancements of up to 2.15\% and an average gain of 0.88\%. To further verify the efficacy of our approach, we applied \scheme\ on the ImageNet dataset. The final two columns of Table~\ref{tbl-acc-merged} reveal that \scheme\ preserved the accuracy of all models at levels nearly equivalent to the baseline (BP), with a negligible average reduction of 0.3\%. In certain cases, such as with DenseNet161 and DenseNet201 our proposed method even increased accuracy by 0.28\% and 0.04\% respectively, and in VGG13, the accuracy remained unchanged.

\subsubsection{Predictor Accuracy}
\label{sec-predictor-accuracy}

\begin{figure}[htpb]
\centering
\begin{subfigure}{0.5\columnwidth}
\centering
\includegraphics[width=\columnwidth]{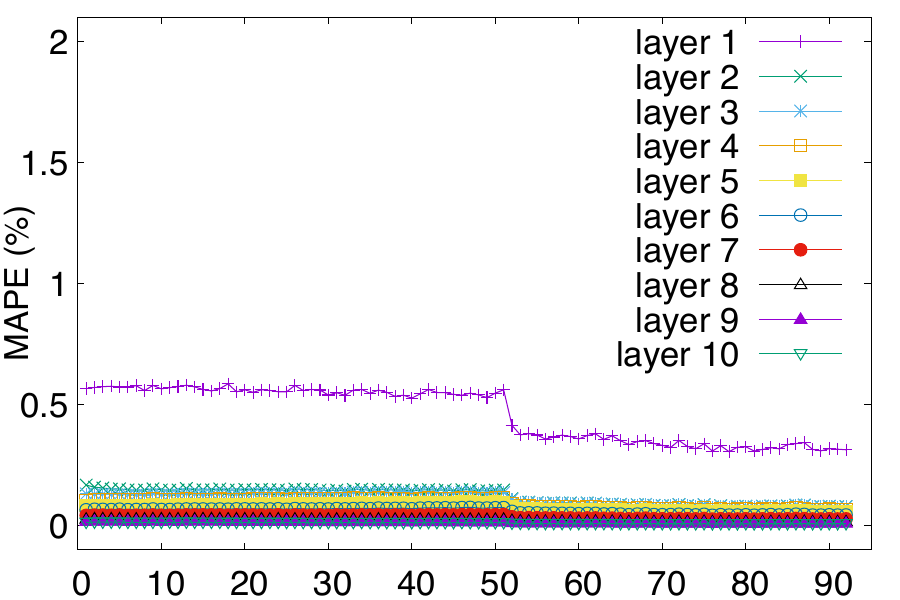}
 \caption{MAPE}
\label{fig-mape}
\end{subfigure}
\hfill
\begin{subfigure}{0.49\columnwidth}
\centering
\includegraphics[width=\columnwidth]{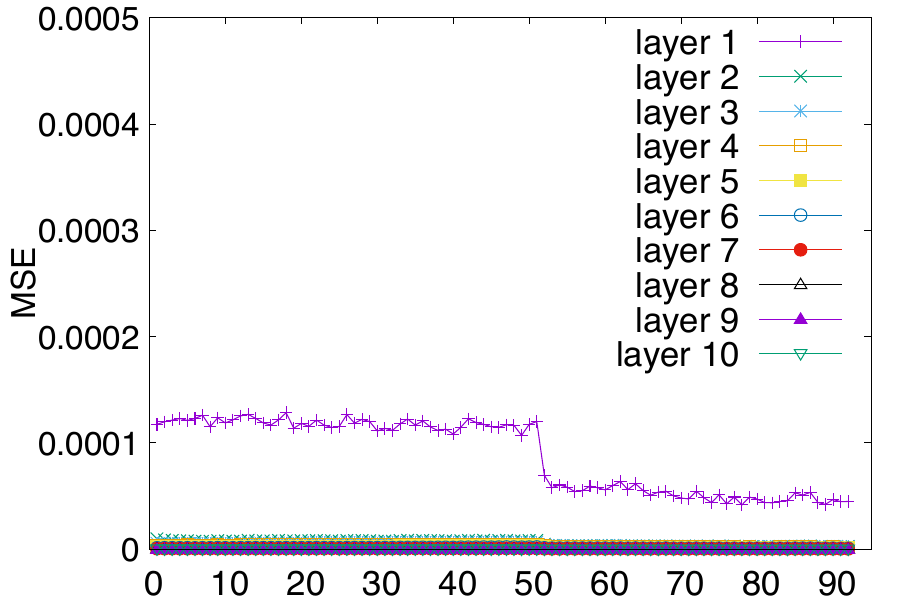}
\caption{MSE}
\label{fig-mse}
\end{subfigure}
\caption{MAPE and MSE of the predictor model in different layers of VGG13 during the training.}
\label{fig-mape-mse}
\end{figure}

We use the Mean Absolute Percentage Error (MAPE) 
to show the accuracy and performance of the predictor model ~\cite{mape}. The accuracy is generally calculated as a ratio, as outlined by the following equation~\cite{mape}:
\begin{equation}
MAPE =\frac{1}{n}\Sigma_{t=1}^{n} \lvert\frac{A_t-P_t}{A_t}\rvert
\end{equation}
Here, $A_t$ is the actual value, $P_t$ is the predicted value, and $n$ is the total number of predicted values. Figure~\ref{fig-mape} shows the MAPE for different layers of the VGG13. The figure illustrates that the MAPE value is below 0.16\% for layers 2-10. It also shows a consistent improvement during the training epochs. For layer 1, the MAPE starts at 0.56\% in the first epoch and decreases to 0.31\% after 90 epochs.
Figure \ref{fig-mse} shows the Mean Squared Error (MSE) of the predictor model during 
training. MSE show similar trends as MAPE.

\subsection{Case Study: VGG13}
\label{detail-analysis}
\begin{figure}[htpb]
\centering
\includegraphics[width=0.8\columnwidth]{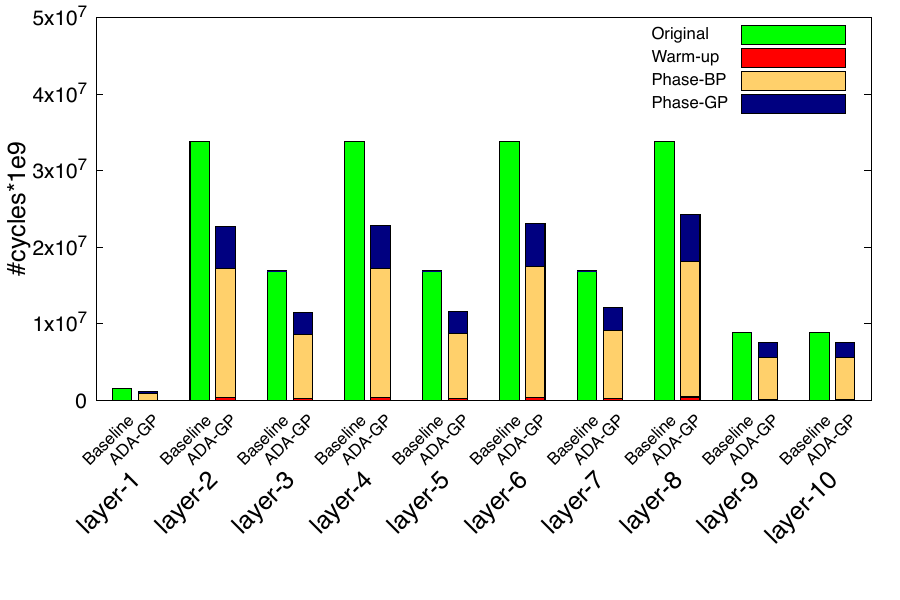}
\caption{Characterization of \scheme\ in VGG13.}
\label{vgg13-cycle-detail}
\end{figure}

We perform an in-depth analysis of VGG13, decomposing the training costs across various layers by employing the \scheme-Efficient approach as well as the conventional BP technique. The outcomes can be seen in Figure~\ref{vgg13-cycle-detail}. In the \scheme-Efficient, we divide the costs into three parts, each corresponding to distinct stages of the training process such as Warm-up (step 1 + step 2), \phasebp, and \phasegp. 

\subsection{Performance Analysis}
\label{experimental-scenarios}
In the \scheme-MAX approach, during \phasebp, the forward pass (FW) of the \synmodel\ model can be computed simultaneously with the FW of the subsequent layer, as well as the backward pass (BW) of the \synmodel\ model alongside the BW of that layer. This method allows us to nearly eliminate the \synmodel\ model's operation in \phasegp, but we must still determine the maximum between the original and \synmodel\ models.
It is essential to wait for the current layer to complete its operation before proceeding to the next layer to avoid conflicts between the operations of different layers.

In the \scheme-Efficient method, the \synmodel's weights are constantly stored in designated memory; however, there is no additional processing element (PE) to perform the \synmodel\ model's operations in parallel with the original model's operations. Consequently, the cost of each layer equals the sum of the original model and \synmodel\ model costs in distinct phases. Similar to the \scheme-MAX approach, the operations between layers are synchronized, initiating the subsequent layer's operation only after the current layer's operation is completed.

In the \scheme-LOW approach, there is no additional memory allocated for the \synmodel\ model weights, requiring us to load them after each original layer operation. Consequently, the expense associated with each layer should encompass the loading of \synmodel\ model weights and the storage of computed results. Nonetheless, following the loading process, the number of operations would be akin to the \scheme-Efficient approach. 

\begin{figure*}[!h]
\centering
    \begin{subfigure}{0.33\textwidth}
    \centering
    \includegraphics[width=\columnwidth]{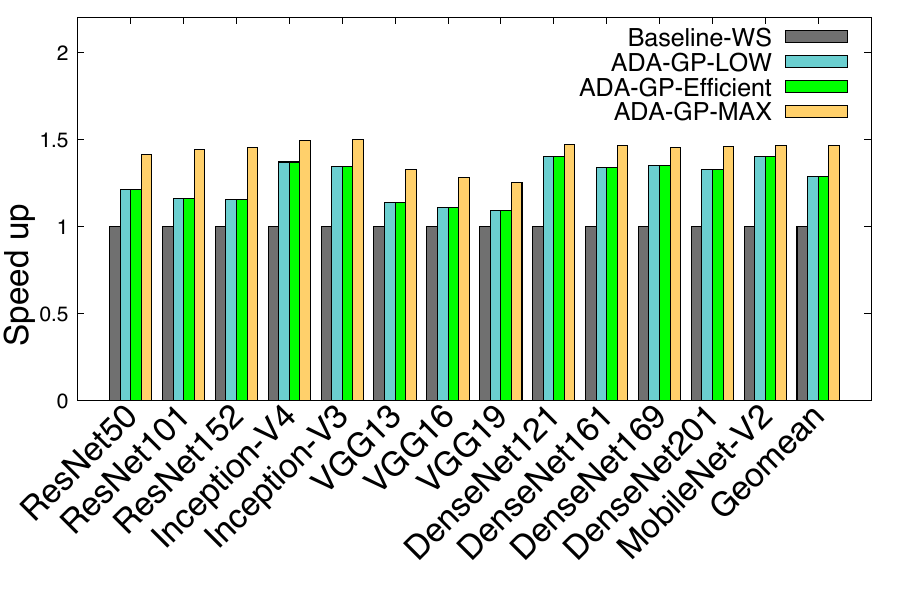}
    \caption{Cifar10 dataset}
    \label{ovr-WS-cifar10}
    \end{subfigure}
    \begin{subfigure}{0.33\textwidth}
    \centering
    \includegraphics[width=\columnwidth]{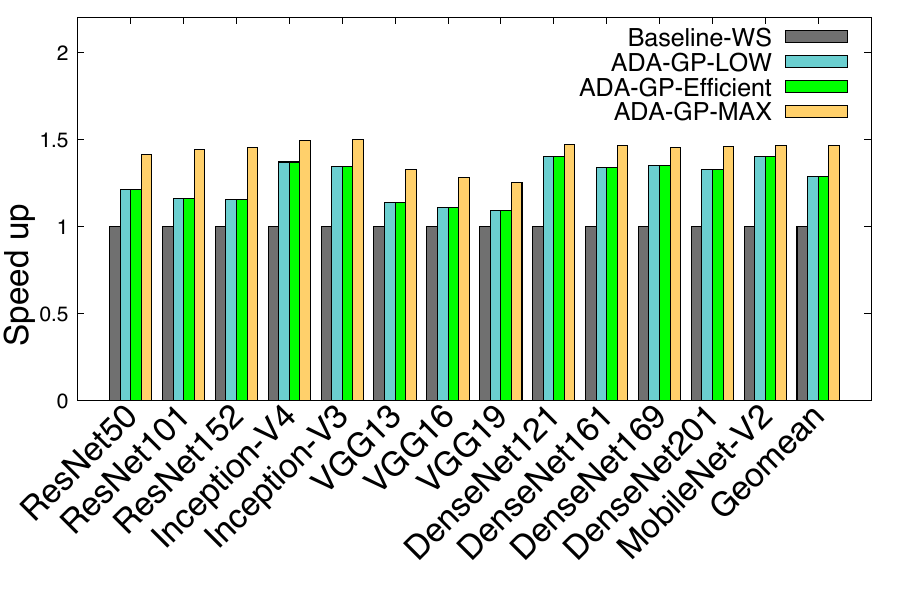}
    \caption{Cifar100 dataset}
    \label{ovr-WS-cifar100}
    \end{subfigure}
    \begin{subfigure}{0.33\textwidth}
    \centering
    \includegraphics[width=\columnwidth]{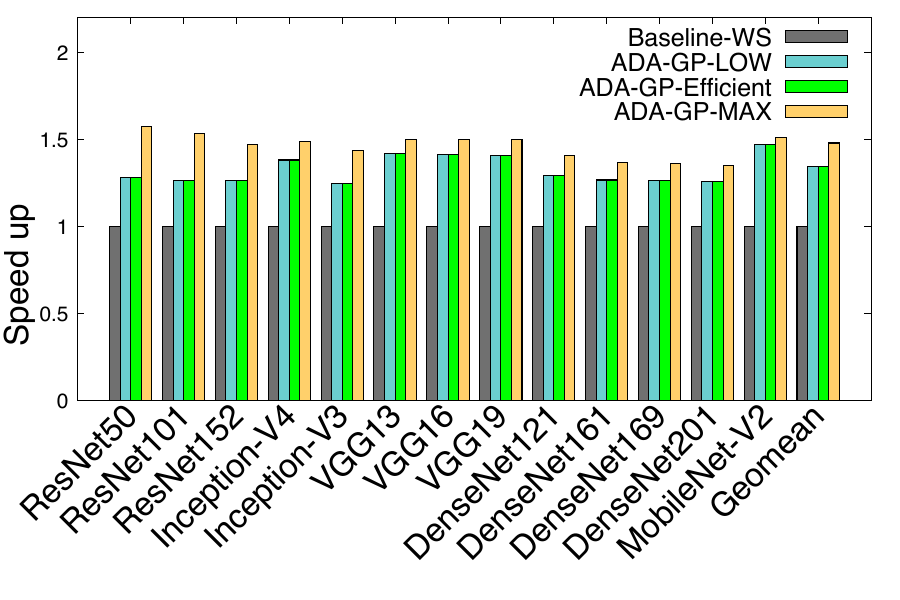}
    \caption{ImageNet dataset}
    \label{ovr-WS-imagenet}
    \end{subfigure}
    \caption{Speed up of \scheme\ over the baseline (BP) in Weight-Stationary (WS) dataflow.} 

    \label{ovr-speedup-WS}
\end{figure*}

Figures~\ref{ovr-WS-cifar10}, \ref{ovr-WS-cifar100}, and~\ref{ovr-WS-imagenet} display the overall acceleration of \scheme-LOW, \scheme-Efficient and \scheme-MAX in comparison to the baseline system. In these figures, the baseline system represents a standard BP process utilizing the Weight-Stationary (WS) dataflow. The performance metrics are reported in relation to the dataset, as the model's structure exhibits slight changes depending on the input size in different datasets. 
As demonstrated in these figures, \scheme-MAX can enhance the training by up to $1.51\times$, $1.51\times$, and $1.58\times$ for the Cifar10, Cifar100, and ImageNet datasets, respectively. Furthermore, it expedites the process by an average of $1.46\times$, $1.46\times$, and $1.48\times$ across all models for the Cifar10, Cifar100, and ImageNet datasets, respectively.

\begin{figure*}[!h]
\centering
    \begin{subfigure}{0.33\textwidth}
    \centering
    \includegraphics[width=\columnwidth]{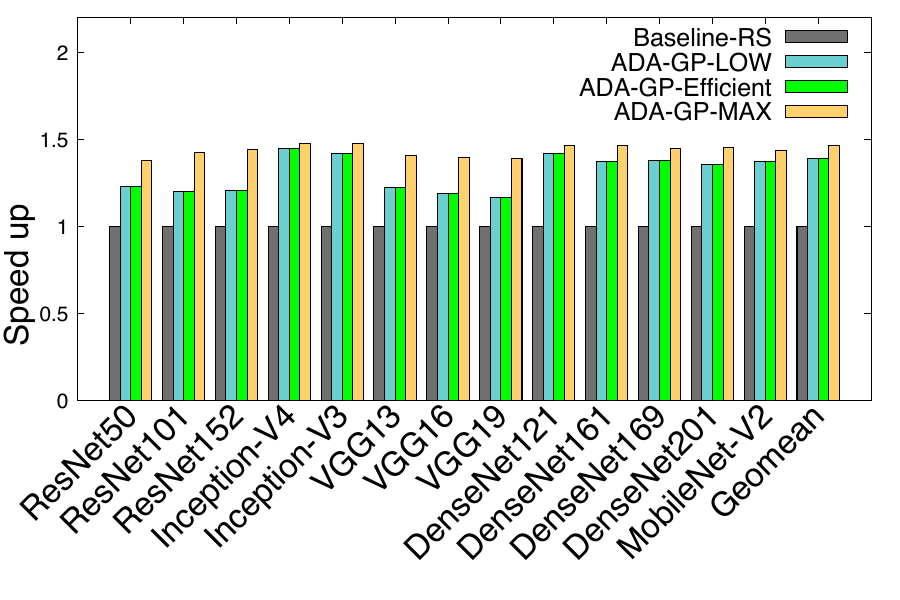}
    \caption{Cifar10 dataset}
    \label{ovr-RS-cifar10}
    \end{subfigure}
    \begin{subfigure}{0.33\textwidth}
    \centering
    \includegraphics[width=\columnwidth]{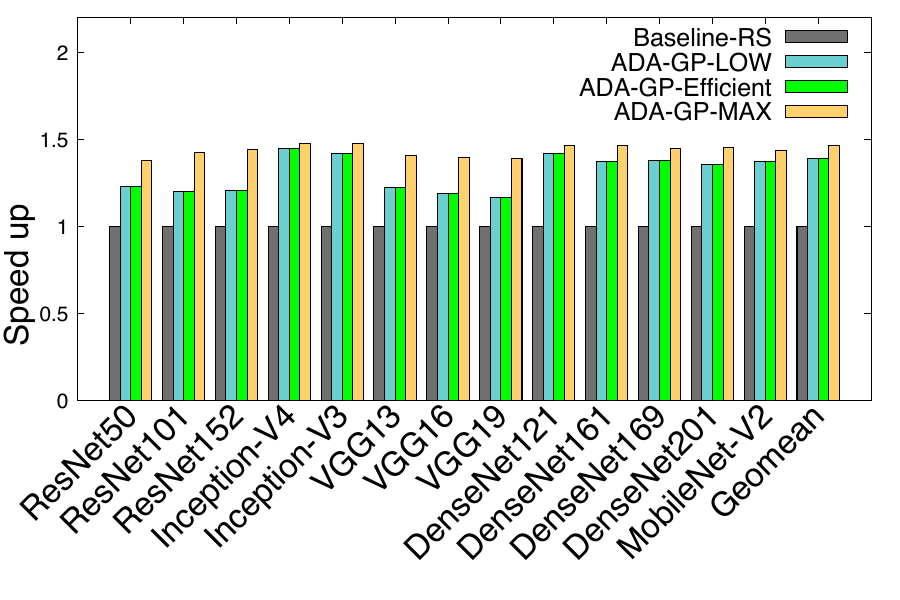}
    \caption{Cifar100 dataset}
    \label{ovr-RS-cifar100}
    \end{subfigure}
    \begin{subfigure}{0.33\textwidth}
    \centering
    \includegraphics[width=\columnwidth]{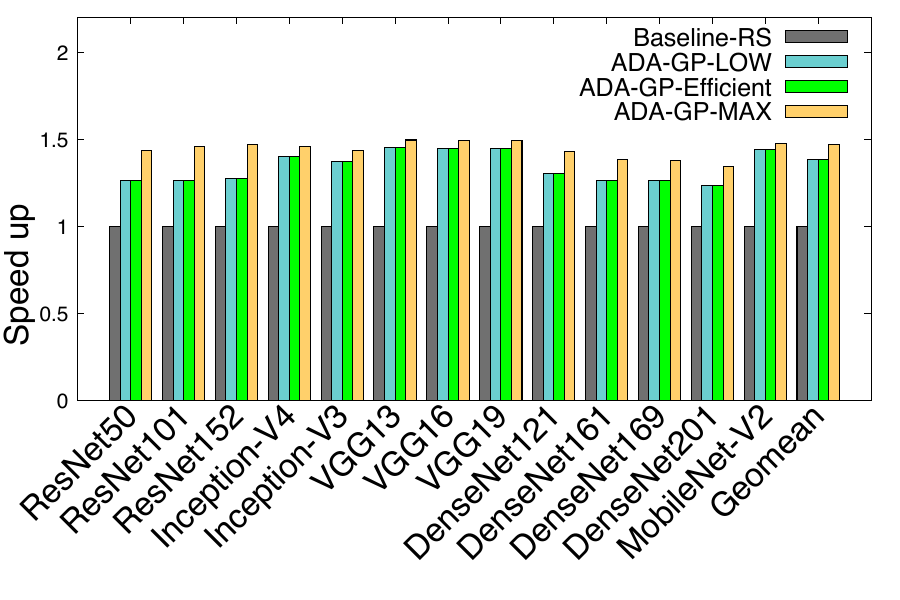}
    \caption{ImageNet dataset}
    \label{ovr-RS-imagenet}
    \end{subfigure}
    \caption{Speed up of \scheme\ over the baseline (BP) in Row-Stationary (RS) dataflow.} 
    

    \label{ovr-speedup-RS}
\end{figure*}

We also perform analogous experiments for the Row Stationary (RS) dataflow. 
Figure \ref{ovr-speedup-RS} illustrates the overall acceleration of \scheme-LOW, \scheme-Efficient, and \scheme-MAX compared to the RS baseline. Figures \ref{ovr-RS-cifar10}, \ref{ovr-RS-cifar100}, and \ref{ovr-RS-imagenet} indicate that \scheme-MAX can boost the training process by up to $1.48\times$ for each of the Cifar10, Cifar100, and ImageNet datasets. Additionally, it increases the training speed on average by $1.46\times$ across Cifar10 and Cifar100 datasets respectively, and $1.47\times$ in the ImageNet dataset.

\begin{figure*}[!h]
\centering
    \begin{subfigure}{0.33\textwidth}
    \centering
    \includegraphics[width=\columnwidth]{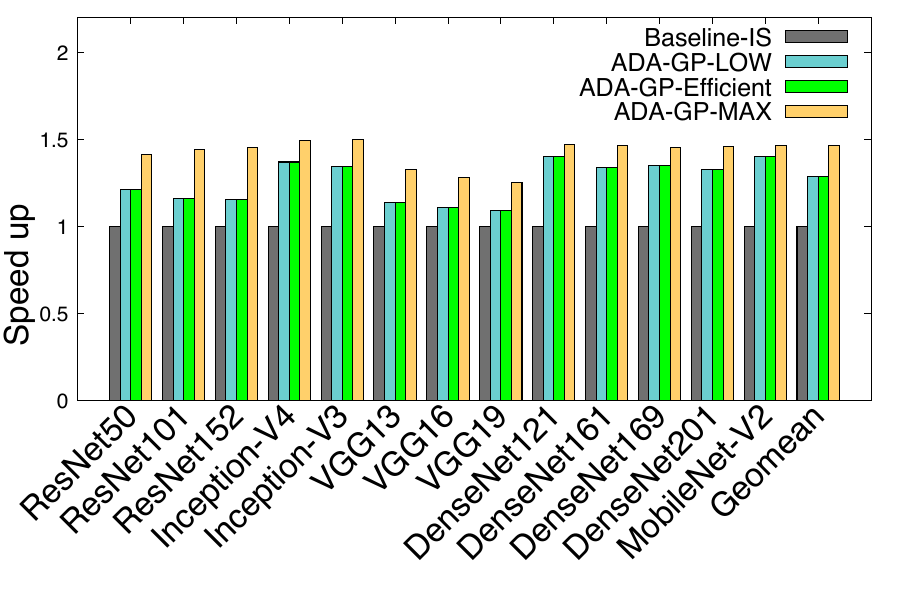}
    \caption{Cifar10 dataset}
    \label{ovr-IS-cifar10}
    \end{subfigure}
    \begin{subfigure}{0.33\textwidth}
    \centering
    \includegraphics[width=\columnwidth]{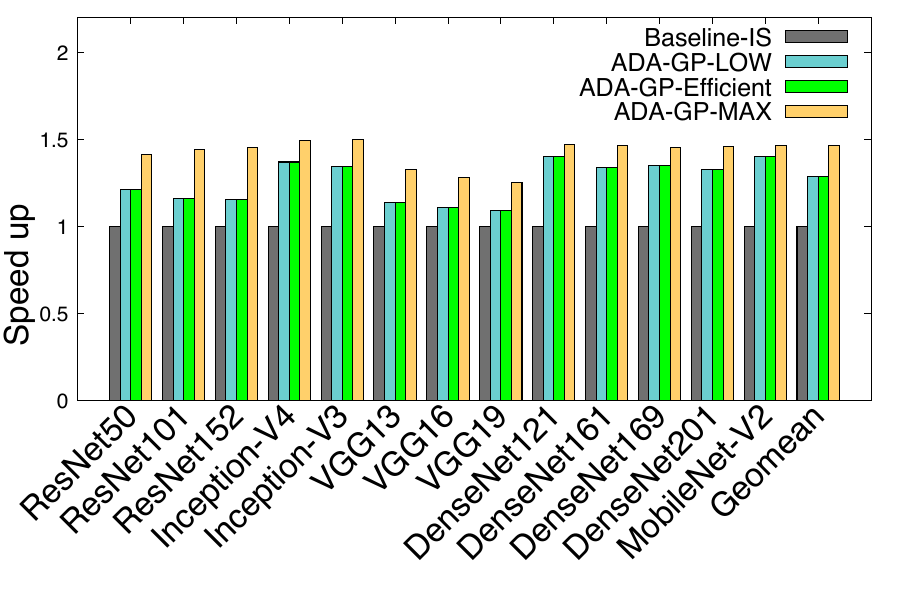}
    \caption{Cifar100 dataset}
    \label{ovr-IS-cifar100}
    \end{subfigure}
    \begin{subfigure}{0.33\textwidth}
    \centering
    \includegraphics[width=\columnwidth]{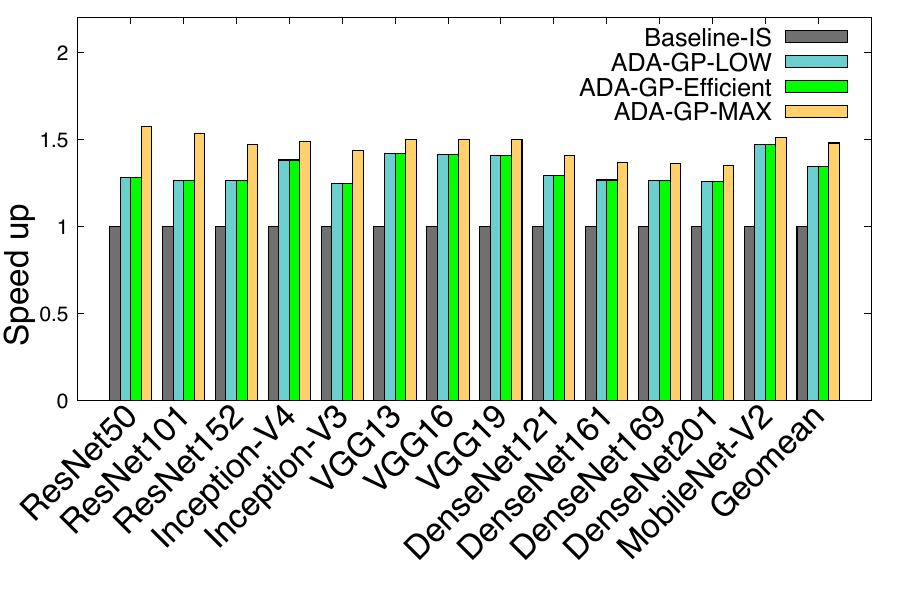}
    \caption{ImageNet dataset}
    \label{ovr-IS-imagenet}
    \end{subfigure}
    \caption{Speed up of \scheme\ over the baseline (BP) in Input-Stationary (IS) dataflow.} 
    

    \label{ovr-speedup-IS}
\end{figure*}

In a similar vein, we carried out additional experiments to demonstrate the acceleration of \scheme-LOW, \scheme-Efficient, and \scheme-MAX over the Input-Stationary (IS) dataflow baseline. Figure~\ref{ovr-speedup-IS} presents a summary of these experimental results. Figures~\ref{ovr-IS-cifar10}, \ref{ovr-IS-cifar100}, and~\ref{ovr-IS-imagenet} reveal that \scheme-MAX can enhance the training process by an average of $1.46\times$, $1.46\times$, and $1.48\times$ for the Cifar10, Cifar100, and ImageNet datasets, respectively. 

\subsection{Evaluation with Transformer and Object Detection Models}
\label{transformer-analysis}

In this section, we employed the \scheme\ technique on Transformer, consisting of three encoding and decoding layers, as well as YOLO-v3 \cite{redmon2018yolov3} object detection models.
We distinguish these models from other deep learning models due to the different employed datasets. 

To assess our approach, for the Transformer model, we utilized the Multi30k \cite{elliott-etal-2016-multi30k} English-German translation dataset.
Table \ref{tbl-per-transformer} shows the overall accuracy and performance comparison of \scheme\ with \aj{the} baseline (BP) design.
\begin{table}[!htpb]
    \centering
    \caption{Accuracy and performance comparison between \scheme\ and \aj{the} Baseline (BP) for Multi30k dataset.}
    \label{tbl-per-transformer}
    \scalebox{0.85}{
    \def\arraystretch{1.3}
    \begin{tabular}{||l||c|c|c|c||} 
    \hline 

     & Val Acc. &  loss & BLEU Score & \#Cycles($\times10^9$) \\
    \hline\hline
	Baseline(BP)	 & 52.42  &	1.61  & 33.52  & 1245.87 \\\hline
	\textbf{\scheme} & 52.14  & 1.65  & 33.4   & 1104.31 \\\hline
 
    \end{tabular}}
\end{table}
As demonstrated in Table~\ref{tbl-per-transformer}, \scheme\ accelerates the training process of the Transformer by a factor of $1.13\times$. Furthermore, \scheme\ does not adversely impact the model's performance and maintains the high accuracy of the Transformer model, achieving nearly identical BLEU Score~\cite{papineni2002bleu} results.

Furthermore, we employed the \scheme\ technique on the YOLO-v3 \cite{redmon2018yolov3} object detection model.
Here, we utilized the Pascalvoc \cite{Everingham15} Visual Object Classes dataset. We set the initial learning rate to 3e-5, weight decay to 1e-4, and IOU threshold to 0.5.  
Table \ref{tbl-per-yolo} shows the overall performance comparison of \scheme\ with \aj{the} baseline (BP) design. 
\begin{table}[!htpb]
    \centering
    \caption{Accuracy and performance comparison between \scheme\ and \aj{the} Baseline (BP) for Pascalvoc dataset.}
    \label{tbl-per-yolo}
    \scalebox{0.85}{
    \def\arraystretch{1.3}
    \begin{tabular}{||l||c|c|c|c||} 
    \hline 
                               &  Class Acc    & Test MAP   & \#Cycles($\times10^9$) \\
    \hline\hline
    Baseline(BP)               & 83.03         & 0.4685     & 11490248 \\\hline
    \textbf{\scheme-Efficient} & 82.51         & 0.4674     & 9846579 \\\hline
    \textbf{\scheme-MAX}       & 82.51         & 0.4674     & 9134071 \\\hline
    \end{tabular}}
\end{table}
As demonstrated in Table~\ref{tbl-per-transformer}, \scheme\ accelerates the training process of the YOLO-v3 by a factor of $1.17\times$ and $1.26\times$ in \scheme-Efficient and \scheme-MAX, respectively. Furthermore, \scheme\ keeps the class accuracy high and achieves nearly identical Test MAP results.

\subsection{Multi-Device Comparative Analysis}
\label{multi-device-analysis}
In this section, we evaluate the performance of the proposed \scheme\ relative to different baseline pipelining techniques using the ImageNet dataset in the context of multi-device hardware systems including GPipe~\cite{huang2019gpipe}, DAPPLE~\cite{fan2021dapple}, and Chimera ~\cite{li2021chimera}. 
We employ the scenario outlined in section \ref{Multi-Device-scenario} to compute the training acceleration. 
We consider a setup with four devices operating concurrently, where each mini-batch is split into four portions (macro-batch), and each device processes one macro-batch at a time. Similarly, we can implement \scheme\ across diverse multi-device hardware systems with varying numbers of devices, offering additional savings on top of their existing configurations. In this section, the duration of each time step is equivalent to the delay of the FW process in a single device for one macro-batch. Throughout the remainder of this section, we will employ this definition of a $step$ in our explanations.

\subsubsection{Comparison with GPipe }

\begin{figure*}[!h]
\centering
    \begin{subfigure}{0.33\textwidth}
    \centering
    \includegraphics[width=\columnwidth]{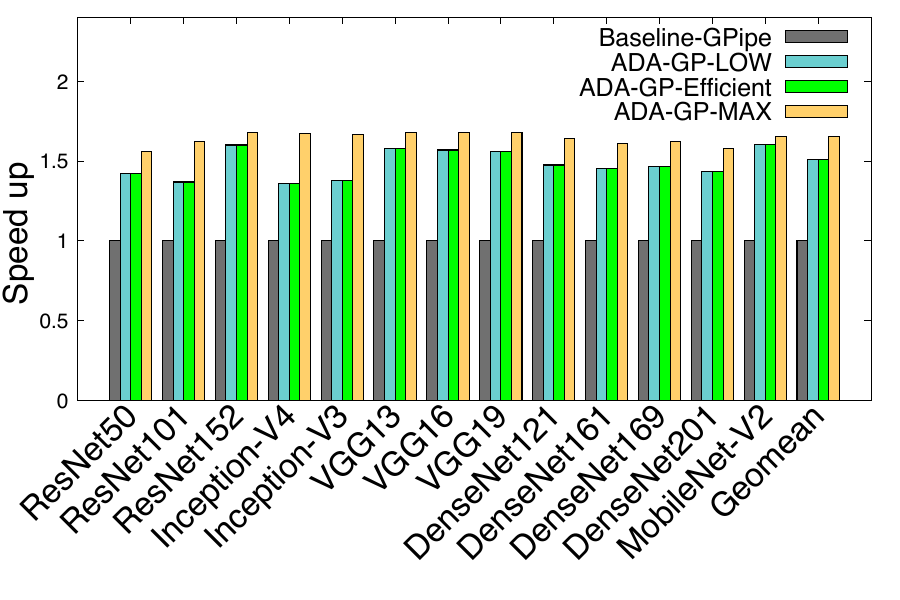}
    \caption{Speed up over GPipe~\cite{huang2019gpipe}.}
    \label{speedup-gpipe}
    \end{subfigure}
    \begin{subfigure}{0.33\textwidth}
    \centering
    \includegraphics[width=\columnwidth]{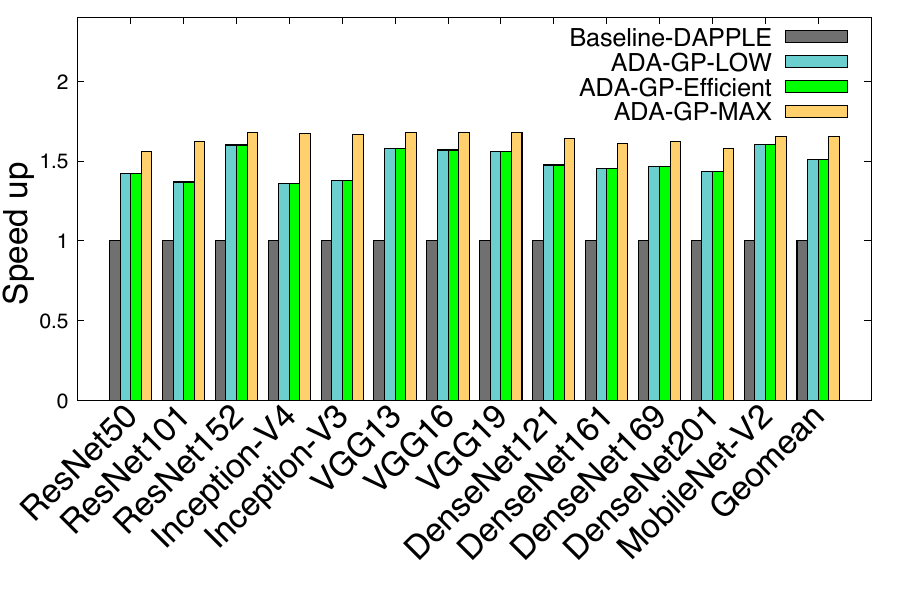}
    \caption{Speed up over DAPPLE~\cite{fan2021dapple}.}
    \label{speedup-dapple}
    \end{subfigure}
    \begin{subfigure}{0.33\textwidth}
    \centering
    \includegraphics[width=\columnwidth]{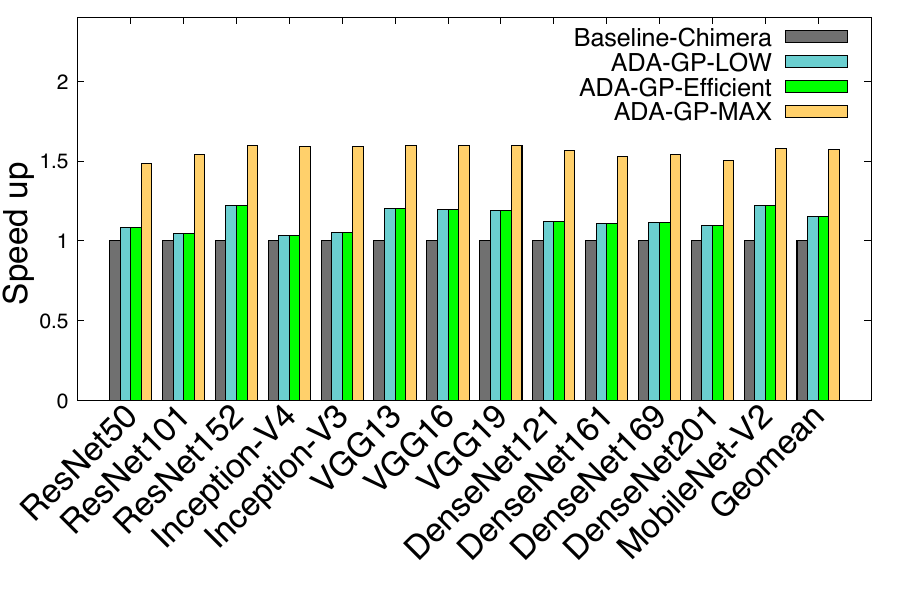}
    \caption{Speed up over Chimera~\cite{li2021chimera}.}
    \label{speedup-chimera}
    \end{subfigure}
    \caption{Speed up of \scheme\ over the baseline pipelining techniques a) GPipe~\cite{huang2019gpipe}, b) DAPPLE~\cite{fan2021dapple}, and c) Chimera~\cite{li2021chimera}.}
    \label{ovr-speedup-WS}
\end{figure*}

As depicted in Figure \ref{ADA-GP-GPipe}, the standard GPipe method takes 21 steps 
to complete the training of $one$ batch. \scheme\ can significantly reduce computations in \phasegp\ by eliminating the conventional backpropagation process. Also, when transitioning from \phasegp\ to \phasebp, \scheme\ only requires 25 steps 
to finish the training of $two$ batches. 
Figure~\ref{speedup-gpipe} depicts the overall acceleration of \scheme\ in comparison to the baseline GPipe method.
%
As seen in Figure~\ref{speedup-gpipe}, \scheme\ accelerates the training process for all deep learning models, achieving up to $1.68\times$ speedup and an average of $1.654\times$ improvement.

\subsubsection{Comparison with DAPPLE }

As illustrated in Figure~\ref{ADA-GP-Dapple}, the DAPPLE method~\cite{fan2021dapple}, similar to GPipe technique,
requires 21 steps 
to complete the training of one batch. The timing of the \scheme\ when applied to DAPPLE also resembles that of the GPipe technique, taking into account the fact that the delay is associated with the DAPPLE design. Figure~\ref{speedup-dapple} demonstrates the extent of \scheme\ acceleration for various deep learning models compared to the baseline DAPPLE design, achieving a maximum speedup of $1.68\times$ and an average improvement of $1.654\times$.


\subsubsection{Comparison with Chimera }
The cutting-edge Chimera~\cite{li2021chimera} technique surpasses all prior methods, potentially addressing numerous gaps in deep learning model training. The Chimera approach manages to complete $one$ batch's training in just 16 steps. 
By incorporating the \scheme\ into the Chimera method, not only do we retain all previous savings during \phasegp, but also when transitioning from \phasegp\ to \phasebp, the scheme necessitates merely 20 steps 
to finish training $two$ batches. Figure~\ref{speedup-chimera} provides the \scheme\ training acceleration for a range of deep learning models. As a result, the scheme effectively speeds up the Chimera training process for these models by up to $1.6\times$ and, on average, $1.575\times$. 

\subsection{Hardware Analysis}
\label{hardware-analysis}

In this section, we discuss the resource usage and power consumption of \scheme\ for both ASIC and FPGA implementations.
Additionally, we provide a detailed comparison of the energy consumption between \scheme\ and the baseline design. We employed CACTI~\cite{cacti} to incorporate cache and memory access time, cycle time, area, leakage, and dynamic power model to calculate the design's energy consumption.

\subsubsection{\scheme\ Hardware Implementation Analysis}
\label{hardware-implementation-analysis}

As mentioned in section~\ref{hardware-implementation}, we proposed three unique designs: \scheme-LOW, \scheme-Effective, and \scheme-MAX, with the goal of balancing acceleration levels and hardware resources. In Table~\ref{tbl-fpga-result}, we compare the resource usage and on-chip power consumption between \scheme\ designs and the baseline for the FPGA implementation.


\begin{table}[htpb]
\centering
\caption{a) Resource usage and b) On-chip power consumption (watt) of \scheme\ designs vs baseline design in FPGA implementation.}
\label{tbl-fpga-result}
    \begin{center} 
    (a) Resource Utilization
    \end{center}
    \scalebox{0.7}{
    \begin{tabular}{||l||c|c|c|c|c||} 
     \hline\hline
     & \#CLB & \#CLB & \#Block & \#Block &  \\
     & LUTs & Registers & RAMB36 & RAMB18 & \#DSP48E1s \\
     \hline\hline
    Baseline          & 472004 & 31402 & 1327 & 514 & 166 \\\hline
    \scheme-LOW       & 489286 & 31856 & 1327 & 514 & 166 \\\hline
    \scheme-Effic. & 493171 & 31916 & 2407 & 514 & 166 \\\hline
    \scheme-MAX       & 494080 & 37452 & 2407 & 514 & 246 \\\hline
     \hline
    \end{tabular}}
    \begin{center} 
    (b) On-chip Power Consumption (watt)
    \end{center}
    \scalebox{0.7}{
    \begin{tabular}{||l||c|c|c|c|c|c|c||} 
     \hline\hline
     &  & CLB &  & Block & & & \\
     &  Clocks & Logic & Signals & RAM & DSPs & Static & Total \\
     \hline\hline
    Baseline          & 0.046 & 0.42  & 0.842 & 0.244 & 0.009 & 2.032 & 3.712 \\\hline
    \scheme-LOW       & 0.047 & 0.446 & 0.857 & 0.243 & 0.001 & 2.032 & 3.745 \\\hline
    \scheme-Effic. & 0.052 & 0.421 & 0.852 & 0.339 & 0.001 & 2.06  & 3.844 \\\hline
    \scheme-MAX       & 0.055 & 0.426 & 0.857 & 0.339 & 0.001 & 2.059 & 3.856 \\\hline
     \hline
    \end{tabular}}
\end{table}

As illustrated in Table~\ref{tbl-fpga-result}, the \scheme-LOW, \scheme-Effective, and \scheme-MAX designs result in a power increase of only 0.8\%, 3.5\%, and 3.8\%, respectively. This rise in power consumption is due to the additional hardware incorporated in the various designs. We conducted another experiment with the baseline and ADA-GP-MAX having the same power. This makes a 10\% increase in the number of PEs in the baseline and an average speedup of 4.31\%, 4.3\%, and 4.47\% for Cifar10, Cifar100, and ImageNet datasets.

Table~\ref{tbl-asic-result} contrasts the area and power consumption of the different \scheme\ designs with the baseline in the ASIC implementation. 

\begin{table}[htpb]
\centering
    \caption{a) Area and b) power consumption (watt) of \scheme\ designs vs baseline design in ASIC implementation.}
    \label{tbl-asic-result}
    \begin{center} 
    (a) Area
    \end{center}
    \scalebox{0.7}{
    \begin{tabular}{||l||c|c|c|c|c||} 
     \hline\hline
     &   &   & Net  & Total & Total \\
     & Combinational & Buf/Inv  & Intercon. & Cell & Area \\
     \hline\hline
    Baseline          & 2331250 & 272483 & 436615 & 2546076 & 2982691 \\\hline
    \scheme-LOW       & 2375188 & 277261 & 445371 & 2590583 & 3035954 \\\hline
    \scheme-Effic.    & 2405881 & 275783 & 440031 & 2622858 & 3062890 \\\hline
    \scheme-MAX       & 2512057 & 287076 & 460157 & 2770979 & 3231136 \\\hline
     \hline
    \end{tabular}}
    \begin{center} 
    (b) Power ($\mu$ watt)
    \end{center}
    \scalebox{0.7}{
    \begin{tabular}{||l||c|c|c|c||} 
     \hline\hline
     &  Internal &  Switching & Leakage & Total \\
     \hline\hline
    Baseline          & 2.26E+04 & 1.72E+03 & 1.99E+05 & 2.24E+05 \\\hline
    \scheme-LOW       & 2.25E+04 & 1.67E+03 & 2.02E+05 & 2.26E+05 \\\hline
    \scheme-Effic.    & 2.27E+04 & 1.80E+03 & 2.00E+05 & 2.25E+05 \\\hline
    \scheme-MAX       & 2.80E+04 & 2.42E+03 & 2.23E+05 & 2.54E+05 \\\hline
     \hline
    \end{tabular}}
\end{table}

As depicted in Table~\ref{tbl-asic-result}, the \scheme-LOW, \scheme-Efficient, and \scheme-MAX designs lead to an increase in the final design area by 1.7\%, 2.6\%, and 8.3\%, respectively. This also results in a rise in the design power. Similar to FPGA implementation we experimented with the baseline and ADA-GP-MAX having the same area. This results in 11\% additional PE in the baseline and an average speedup of 4.63\%, 4.61\%, and 5.53\% for Cifar10, Cifar100, and ImageNet datasets.

\subsubsection{Energy Consumption Analysis}
\label{energy-consumption-analysis}
\begin{figure}[htpb]
\centering
\includegraphics[width=0.95\columnwidth]{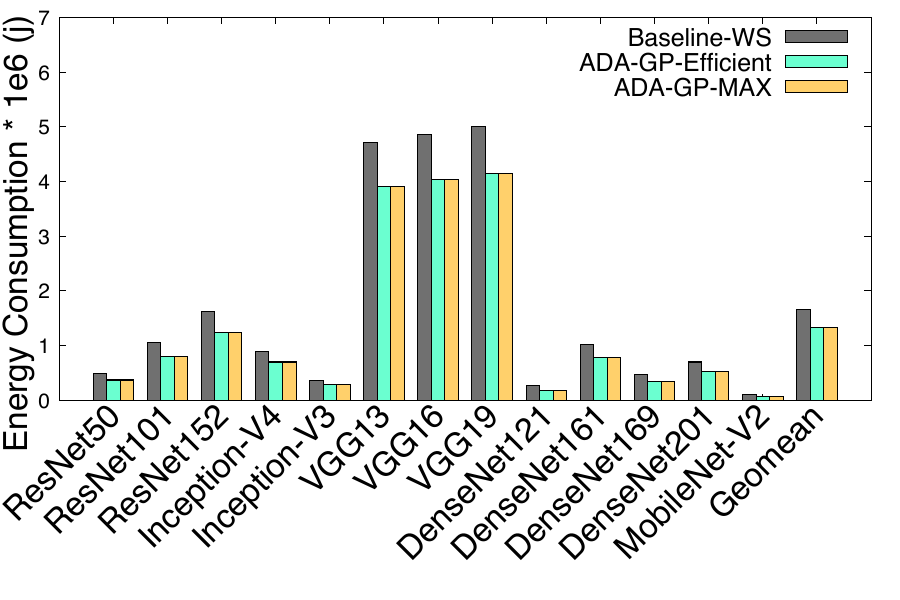}
\caption{The energy consumption comparison between baseline back propagation and \scheme\ designs.}
\label{energy-consumption-cacti}
\end{figure}

In Figure~\ref{energy-consumption-cacti}, the energy consumption associated with memory access during the training process for both the baseline and \scheme~ methods is compared. As a result, \scheme~ enhances energy efficiency for all models, resulting in an average reduction of energy consumption by 34\%.

It is worth mentioning that the presented results do not take into account the savings achieved by reducing the number of synchronization steps, but only reflect the savings from reducing the number of memory read/write operations.

\section{Conclusions}
\label{sec:conclusion}
In this paper, we proposed \scheme, the {\bf first} approach to use gradient prediction to improve the performance of DNN training while maintaining accuracy. \scheme\ warms up the predictor model during the initial few epochs. After that \scheme\ alternates between using backpropagated gradients and predicted gradients for updating weights. As the training proceeds, \scheme\ adaptively decides when and for how long gradient prediction should be used. \scheme\ uses a single predictor model for all layers and uses a {\em novel} tensor reorganization to predict a large number of gradients. We experimented with fifteen
DNN models using three different datasets - Cifar10, Cifar100, and ImageNet. Our results indicate that
ADA-GP can achieve an average speed up of 1.47×
with similar or even higher accuracy than the baseline
models. Moreover, due to the reduced off-chip memory
accesses during the weight updates,
ADA-GP consumes 34\% less energy compared
to the baseline accelerator.



\bibliographystyle{ACM-Reference-Format}
\bibliography{refs}

\end{document}